\pdfoutput=1

\documentclass[11pt]{article}

\usepackage[final]{acl}

\usepackage{times}
\usepackage{latexsym}

\usepackage[T1]{fontenc}

\usepackage[utf8]{inputenc}

\usepackage{microtype}

\usepackage{inconsolata}

\usepackage{hyperref}       %
\usepackage{url}            %
\usepackage{booktabs}       %
\usepackage{amsfonts}       %
\usepackage{nicefrac}       %
\usepackage{microtype}      %
\usepackage{xcolor}         %
\usepackage{color}
\usepackage{colortbl}
\usepackage{multirow}
\usepackage{tcolorbox}
\usepackage{booktabs}
\usepackage{subfig}
\usepackage{wrapfig,lipsum}
\usepackage[russian,english]{babel}
\usepackage{bm, amsfonts, amsmath}

\usepackage{algorithm}
\usepackage{algorithmic}
\newcommand{\LineComment}[1]{\vspace*{0.35em}\small\textit{\# #1}}
\usepackage{amsthm}

\newtheorem{theorem}{Theorem}[section]

\newtheorem{definition}[theorem]{Definition}

\makeatletter
\def\@fnsymbol#1{\ensuremath{\ifcase#1\or \dagger \or  \ddagger\or
   \mathsection\or  \text{*}\or \mathparagraph \or  \| \or **\or \dagger\dagger
   \or \ddagger\ddagger \else\@ctrerr\fi}}
\makeatother
\renewcommand{\thefootnote}{\fnsymbol{footnote}}

\title{Mitigate Position Bias in LLMs via Scaling a Single Hidden States Channel}

\author{Yijiong Yu$^{1}$\footnotemark[1], Huiqiang Jiang$^{2}$, Xufang Luo$^{2}$, Qianhui Wu$^{2}$, Chin-Yew Lin$^{2}$, \\
  \bf  Dongsheng Li$^{2}$, Yuqing Yang$^{2}$, Yongfeng Huang$^{1}$, Lili Qiu$^{2}$\\
  $^{1}$Tsinghua University, $^{2}$Microsoft Corporation\\
  \texttt{yuyj22@mails.tsinghua.edu.cn,yfhuang@tsinghua.edu.cn}\\
 \texttt{\{hjiang,xufluo,qianhuiwu,cyl,dongsli,yuqyang,liliqiu\}@microsoft.com}\\
}

\begin{document}

\maketitle
\footnotetext[1]{Work during internship at Microsoft.}
\renewcommand{\thefootnote}{\arabic{footnote}}  %

\begin{abstract}
Long-context language models (LCLMs) can process long context, but still exhibit position bias, also known as ``lost in the middle'', which indicates placing key information in the middle of the context will significantly affect performance. To mitigating this, we first explore the micro-level manifestations of position bias, concluding that attention weights are a micro-level expression of position bias. Then we identify that, in addition to position embeddings, positional information in hidden states also contributes to position bias, and it manifests itself in specific channels of hidden states, called positional hidden states. Based on these, we propose a method to mitigate position bias by scaling positional hidden states. Experiments on NaturalQuestions Multi-document QA, KV retrieval and LongBench, using various models including RoPE models, context window-extended models, and Alibi models, demonstrate the effectiveness and generalizability of our approach. Our method can improve performance by up to 15.2\% in ``lost in the middle'' benchmark by modifying just one channel of hidden states.
Our code is available at \url{https://aka.ms/PositionalHidden}.

\end{abstract}

\section{Introduction}

Long-context language models (LCLMs)~\citep{reid2024gemini, liu2023world, young2024yi, Abdin2024Phi3TR, deepseekv2} have recently garnered significant attention within the community, enabling LLMs to handle longer and more complex tasks such as long-context question-answering~\citep{caciularu2023peek, li_long-context_2024}.
However, recent research \citep{li_long-context_2024,liu_lost_2023,li_loogle_2023,shi_large_2023,tang_large_2023,junqing2023never,zhang_found_2024} shows that even long-context LLMs often fail to utilize all context information effectively, exhibiting a ``lost in the middle" bias where middle context information is ignored. This issue affects all types of LLMs, regardless of their architecture or size, and worsens with longer contexts.

Previous works have analyzed this issue from the perspectives of data distribution~\citep{junqing2023never,yu_training_2024,an_make_2024} and position embeddings~\citep{zhang_found_2024,chen_fortify_2023}. For example, FILM~\citep{an_make_2024} addresses position bias by constructing data with key information distributed in various positions for supervised fine-tuning (SFT). 
Ms-PoE~\citep{zhang_found_2024} mitigates position bias by interpolating RoPE~\citep{su_roformer_2022} using head-wise scaling factors.

On the other hand, some works~\citep{haviv_transformer_2022,wang_length_2024,chi_latent_2023} have proven that, besides position embeddings, the hidden states of LLMs can also convey positional information, which is generated by the causal attention mask. Although various works (see related works in Appendix \ref{sec:related}) have attempted to mitigate position bias, few of them have related position bias to positional information in hidden states, which may be another non-negligible source of position bias.

To verify the relation between position bias and positional information in hidden states, we conduct a series of experiments. First, as many previous works \cite{yin_stablemask_2024,wang2024eliminatingpositionbiaslanguage,chen_fortify_2023} have observed, we find U-shaped attention weights consistent with position bias in specific layers. Second, we use perturbation experiments to prove position bias is indeed affected by positional information in hidden states. Third, We identify some channels called ``positional channel'' (the hidden states corresponding to them are called ``positional hidden states''), whose values can manifest absolute positional information, through a process of hypothesis and subsequent verification. Finally, we confirm these channels can indeed affect position bias.

Naturally, to mitigate position bias, a direct way is to eliminate (or minimize) its potential cause: positional hidden states. So, we propose a position bias mitigation method named \textbf{``scale positional hidden states''}. Specifically, we first design a heuristic searching algorithm that quickly identifies which channels (along the hidden size axis) of the hidden states are potential positional channel, using monotonicity and smoothness as indicators. Then We select the optimal channel among them based on the loss on a calibration dataset. Next, based on extensive engineering experience, we design a well-designed attention modification algorithm that let the scaled hidden states only influence the attention queried by the last token, to eliminate positional bias in attention, while avoiding interfering too much with the original attention leads to unstable model performance.

Extensive experiments on various models, including LLaMA-2 \citep{touvron2023llama}, Vicuna~\citep{vicuna2023}, Mistral \citep{jiang2023mistral}, Gemma \citep{team2024gemma}, Qwen \citep{qwen}, and MPT \citep{MosaicML2023Introducing}, and across different tasks, including Multi-document QA, KV retrieval, LongBench~\citep{bai2023longbench} benchmark, demonstrate that our method effectively mitigates position bias by modifying only one channel of the hidden states of the model. In addition, we test on timeline reorder task~\citep{li_loogle_2023} and MMLU~\citep{hendrycks_measuring_2021} to show our methods has minimal side effects on the original capabilities of the model.

Our main contributions are as follows:
\begin{enumerate}
    \item We are the first to discover that the positional information in hidden states is also the cause of position bias, especially the bias to the beginning.
    \item We are the first to find explicit, visible hidden states channels which are approximately correlated to absolute token positions, called positional hidden states. And we confirm they can affect position bias.

    \item We propose a method for identifying and scaling the positional hidden states to mitigate position bias.
\end{enumerate}

\section{Positional Information in hidden states affects position bias}
\label{sec:motivation}

\begin{figure}[t]
  \centering
\includegraphics[width=1\columnwidth]{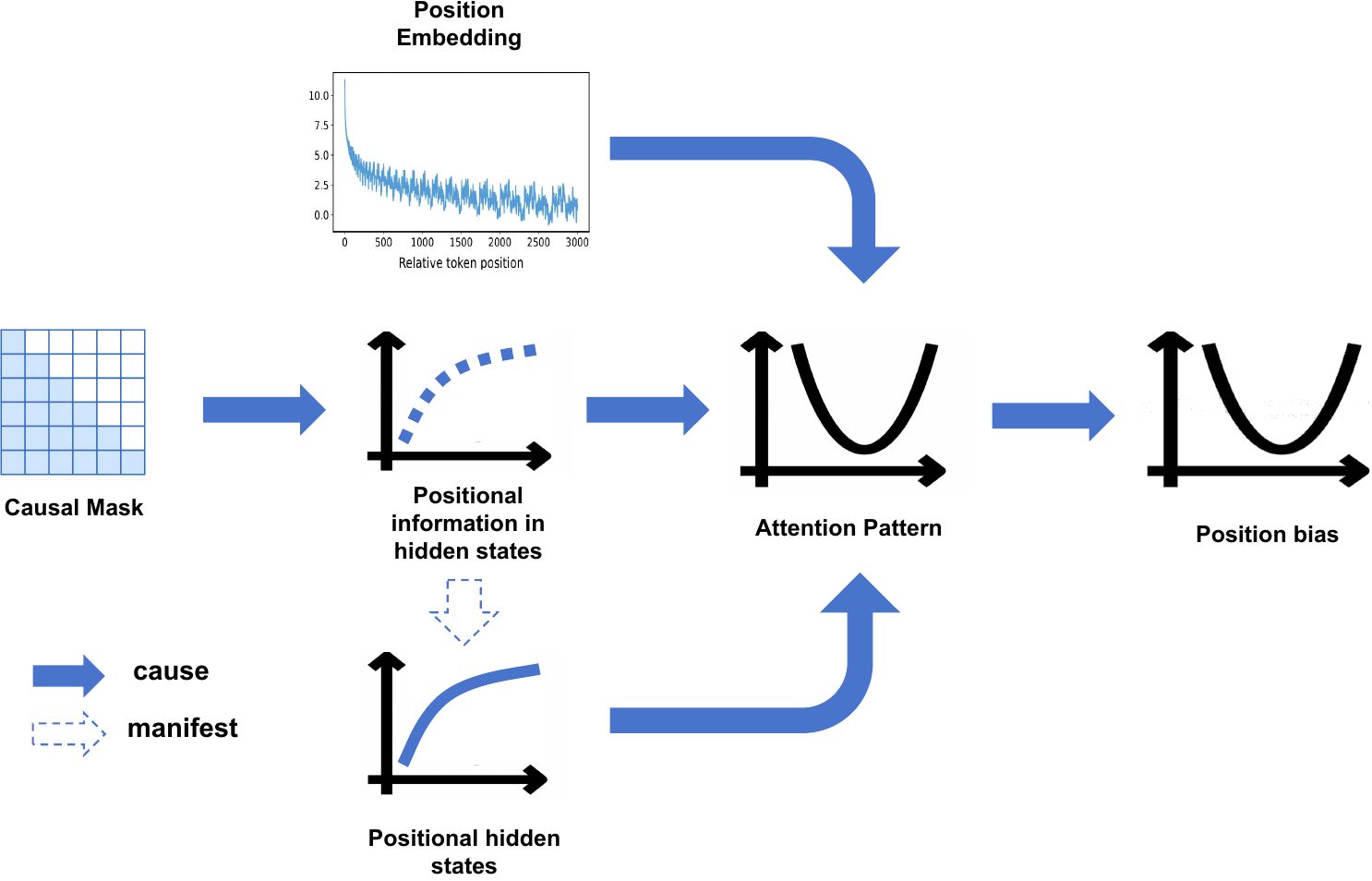}
  \caption{The relationship between causal mask, positional information in hidden states, positional hidden states, position embedding, attention pattern and position bias of model's performance.
  }
  \label{fig: relation}
\end{figure}

In this section, we first identify patterns in attention weights that closely correspond to position bias. Then, through modifying the causal mask in former layers and observing the change of attention in latter layers, we demonstrate position-related attention bias can also be affected by the positional information in hidden states. Third, we extract specific hidden states channels which has monotonicity with the token position, named ``positional hidden states'', and find that they bears responsibility for the emergence of position bias. To sum up, we show our findings about position bias in this section in Figure \ref{fig: relation}.

\subsection{Microscopic Manifestations of Position Bias in Transformers: Attention Weight Patterns}

To explore the micro-level manifestations of position bias in Transformers, we analyze the attention weights in a classic long-context task, KV retrieval, which requires the model to retrieve the value of the given key from a list containing 50 Key-Value pairs (see Appendix~\ref{app:obtain attn} for detailed prompts). The KV pair corresponding to the given key in the question is called the \textbf{Gold KV}. In attention analysis, we average all the attention weights from the last token of the question to the tokens of the $i$-th KV pair as the model's ``attention to the $i$-th KV''. More details about how we calculate attention are in Appendix~\ref{app:obtain attn}.

As shown in Figures~\ref{fig:attn pattern kv}, in deep layers, the model exhibits retrieval-like behavior, focusing on key information, forming a diagonal pattern observed in Figure~\ref{sfig:layer15 attn}. So we call them \textbf{retrieval-related layers}. While in other shallow layers, it always focus most attention on the start or end of the prompt, wherever the key information is located, exhibiting vertical lines patterns, as shown in Figure~\ref{sfig:layer5 attn}. Attention patterns of all the layers are shown in Appendix~\ref{app:Attention Distribution}, based on which we roughly regard that layers 15\textasciitilde20 belong to the retrieval-related layers.

In retrieval-related layers, it can be observed that the attention weights for key information (Gold KV) exhibit patterns similar to position bias: when key information is located at the start or end of the prompt, the attention weights focused on it are relatively higher, while in the middle, lower. Moreover, we extract the attention to key information (average of layers 15\textasciitilde20) with different context length. As shown in Figure \ref{sfig:attention_to_gold_KV}, as the context length grows, the attenuation of attention weights with respect to position becomes more pronounced.

Furthermore, in Appendix~\ref{app:changing_attention}, we find artificially adjusting the attention weights to the key information can directly improve the corresponding retrieval accuracy (this is intuitive). Thus, we confirm that position bias of model performance is to a large extent caused by the bias in attention weight patterns. So in the following experiments, we also use attention weights as the indicator of whether the model has successfully retrieved the key information, as well as the retrieval accuracy.

\begin{figure*}[htb]
  \centering
  \subfloat[Vertical Line Pattern]{
    \label{sfig:layer5 attn}
    \includegraphics[height=0.45\columnwidth]{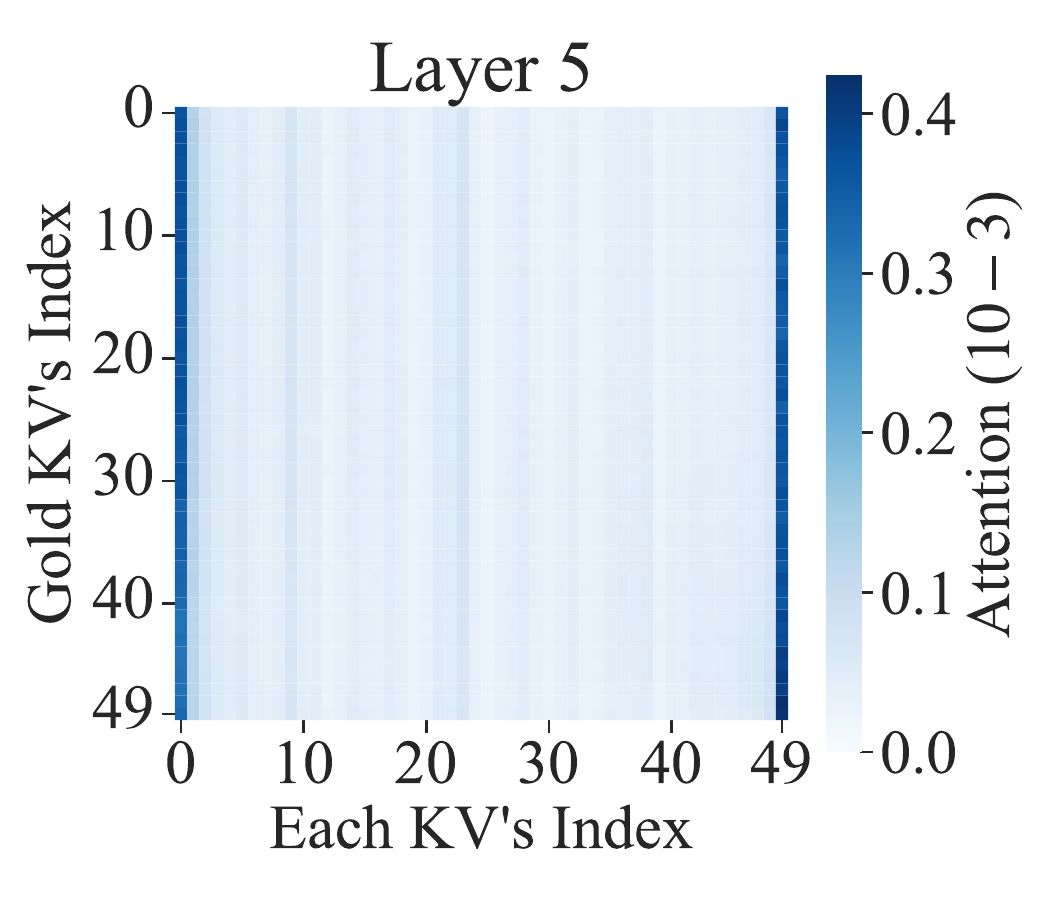}}
    \hspace{0.1em}
      \subfloat[Diagonal Line Pattern]{
    \label{sfig:layer15 attn}
    \includegraphics[height=0.45\columnwidth]{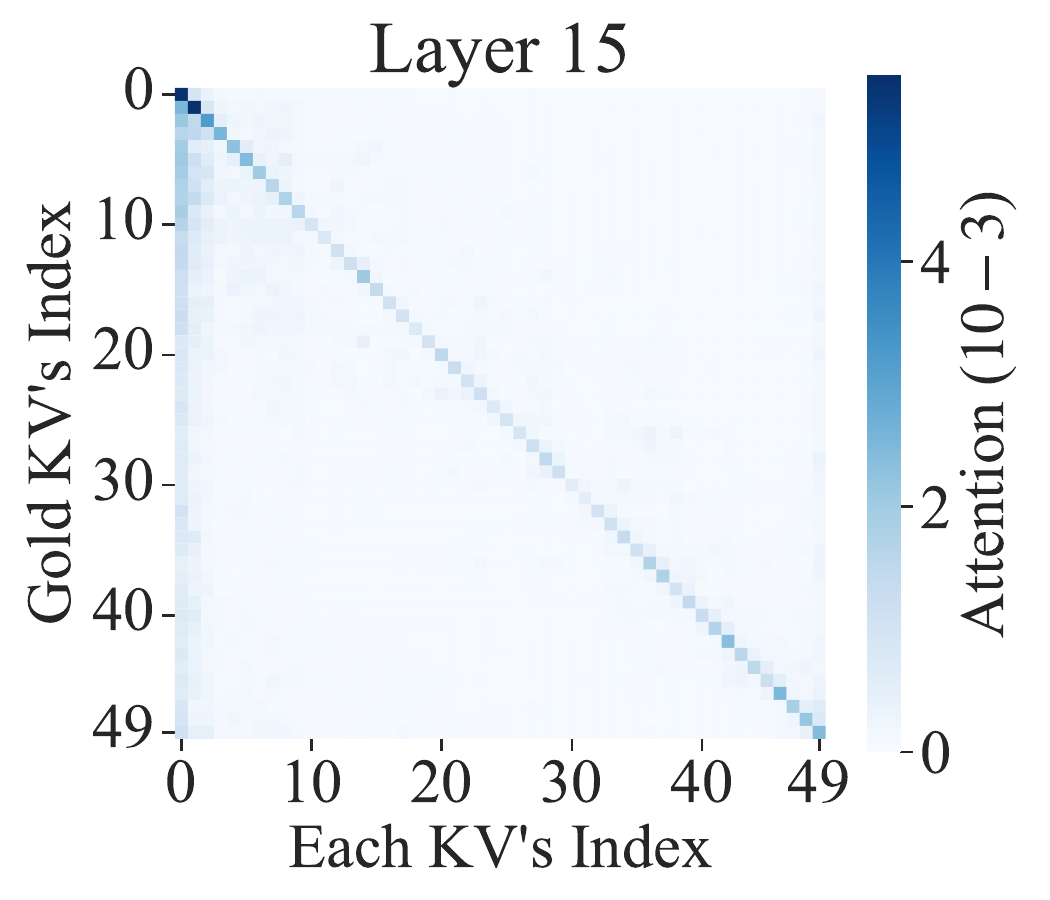}}
    \hspace{0.1em}
  \subfloat[Across Context Lengths]{
    \label{sfig:attention_to_gold_KV}
    \includegraphics[height=0.45\columnwidth]{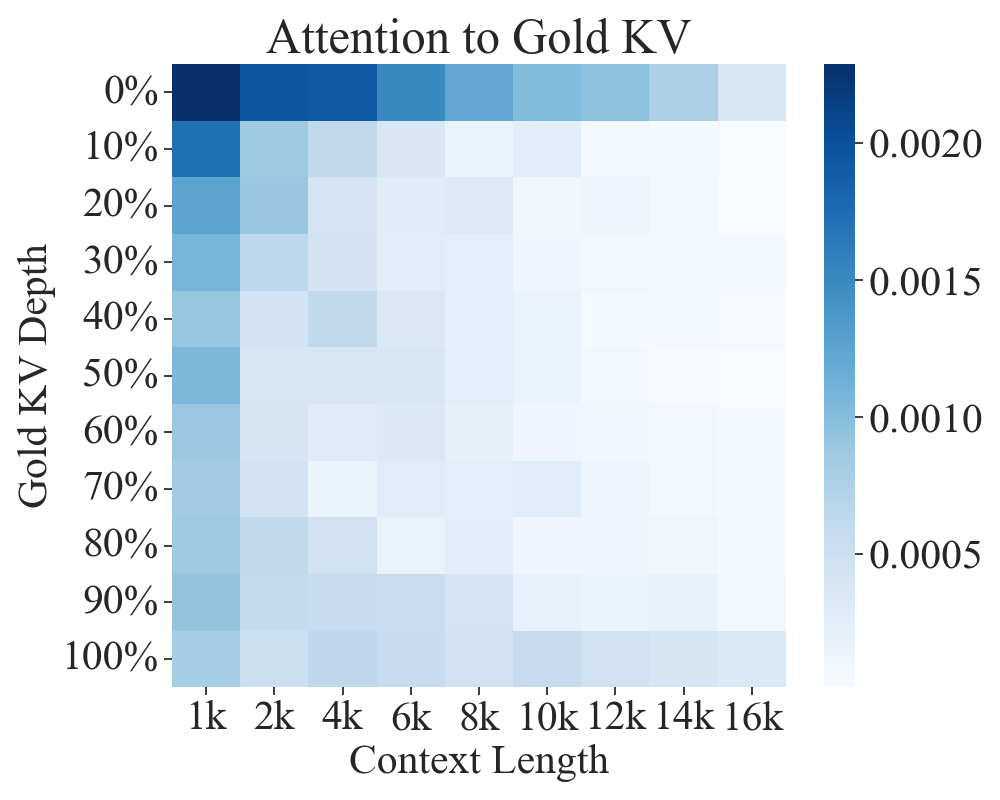}}
  \caption{Attention distribution of the gold KV pair to each KV pair across different positions on the KV retrieval task~\citep{liu_lost_2023} using Mistral-7B~\citep{jiang2023mistral}. (a) and (b) show the results averaged across all heads of the layer. (c) shows the attention of the ground-truth KV to the ground-truth KV (i.e., diagonal lines from (b)) across different context lengths.}
  \label{fig:attn pattern kv}
\end{figure*}

\subsection{Positional Information in Hidden States Also Contributes to Position Bias}
\label{sec:crop mask}

\begin{figure*}[htb]
  \centering
  \includegraphics[width=0.8\linewidth]{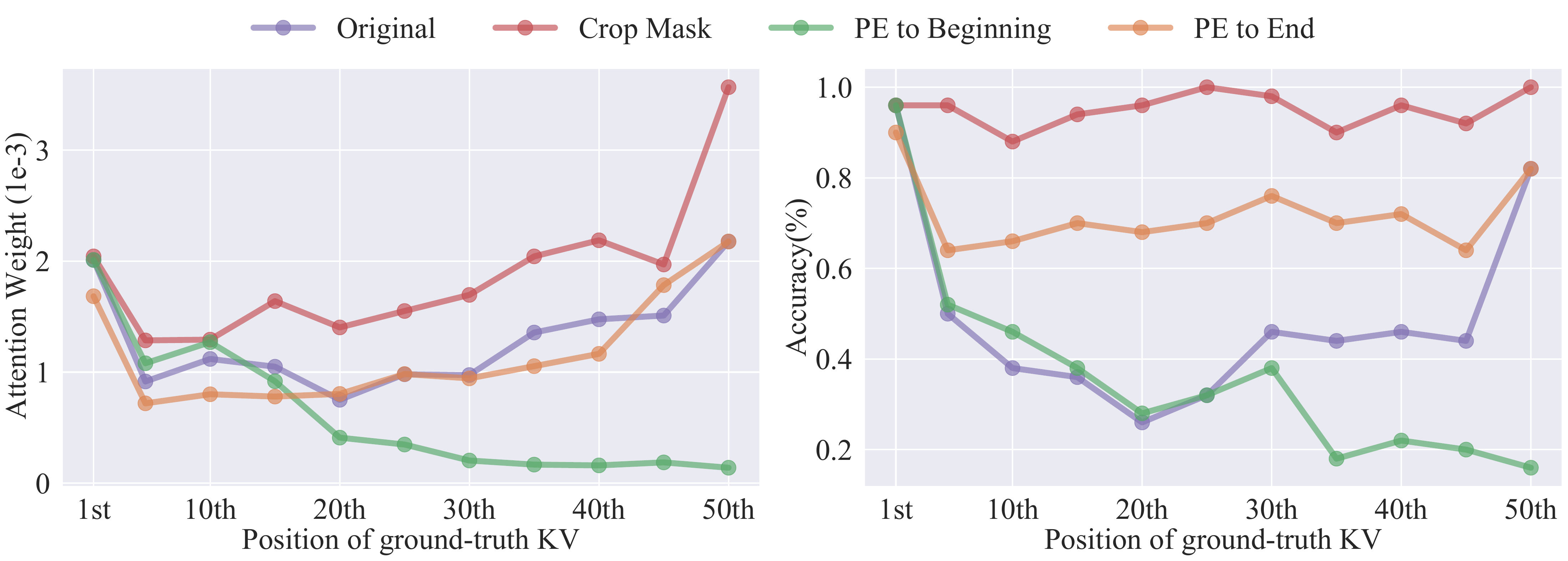}

  \caption{Performance and attention of different methods with the ground-truth KV at different positions in the KV retrieval task~\citep{liu_lost_2023} using Mistral-7B~\citep{jiang2023mistral}.}
  \label{fig:motivations_position_bias}
\end{figure*}

Recent works~\citep{haviv_transformer_2022,wang_length_2024,chi_latent_2023} have indicated that, besides position embeddings, the causal mask can also introduce positional information, which is then stored in the hidden states of LLMs. However, there is no research about whether such information will affect position bias. Therefore, we design a perturbation experiment using Mistral-7B-v0.2 (32 layers) on the classic KV retrieval task, where we modify the causal mask, aiming at changing position information in hidden states, and then observe whether position bias is affected. 

In the experiment, only the causal mask of the layers 2 to 8 are modified, but we mainly observe the attention change in retrieval-related layers, specifically, the average attention of layers 15 to 20 is calculated. This setting avoids modifying the causal mask directly shifts the attention, letting us only see the effect of the positional information in hidden states.

The modification to the causal mask is called \textbf{Crop Mask}, which alters it so the tokens of the gold KV pair can only see itself, but not previous tokens (details about this operation are in Appendix \ref{app:Modify causal mask and PE}). Based on the theory of previous works, we posit this modification will make the positional feature of the tokens of the gold KV pair similar to that of the first KV pair.

In addition, as a comparison, we also shift the position embedding of the tokens of the gold KV (in every layer) as comparisons. Such shift includes: (1) \textbf{PE to Beginning}, which assigns the position IDs of the first KV pair to that of the gold KV pair; (2) \textbf{PE to End}, which assigns the position IDs of the last (the 50th) KV pair to it.

As shown in Figure~\ref{fig:motivations_position_bias}, the original model apparently exhibits a ``lost in the middle'' pattern: when gold KV is in the middle position, the KV retrieval accuracy as well as the attention to it is much lower.

The most notable result is, modifying the casual mask effectively enhances the attention to the gold KV, as well as the retrieval accuracy, whatever its position is. It even lets the attention at the middle be improved to almost on par with the beginning.

As for PE modification, ``PE to end'' has a certain degree of improving attention to the gold KV, but can hardly allow the model's performance to match the accuracy when the gold KV pair is positioned at the start or end of the prompt. In contrast, ``PE to Beginning'' results in a noticeable performance drop as well as attention weight reduction when the gold KV is originally close to the end.  

This phenomenon indicates us that, the direct cause of the attention improvement in retrieval-related layers, can only be the information transmitted through hidden states (because PE and the causal mask in these layers are both not modified). That is, besides position embedding, the positional information in hidden states, which is introduced by causal mask, is also an important factor affecting position bias.

\begin{figure*}[htb]
  \centering
  \vspace{-10pt}
  \subfloat[Mistral-7b]{
    \includegraphics[height=0.5\columnwidth]{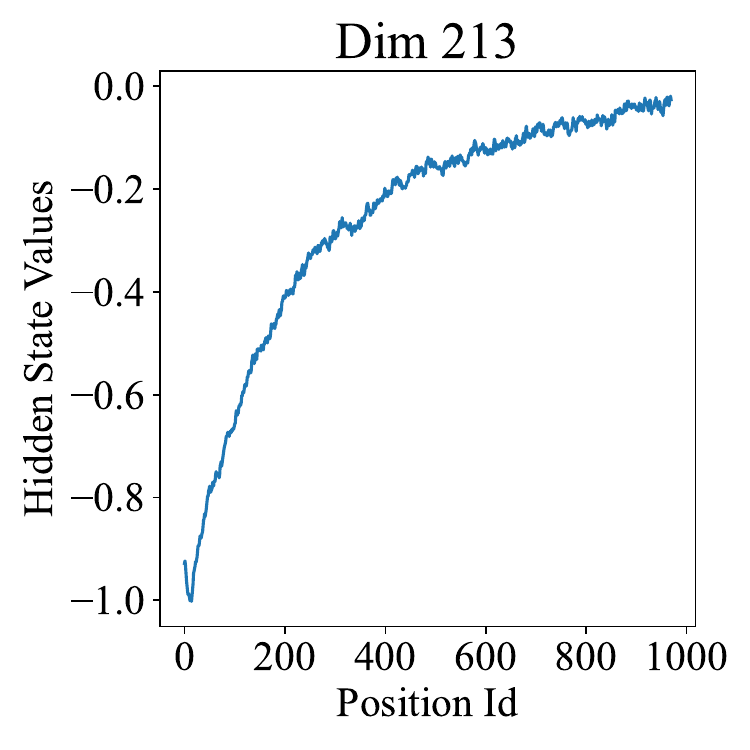}}
  \subfloat[Llama2-7b]{
\includegraphics[height=0.5\columnwidth]{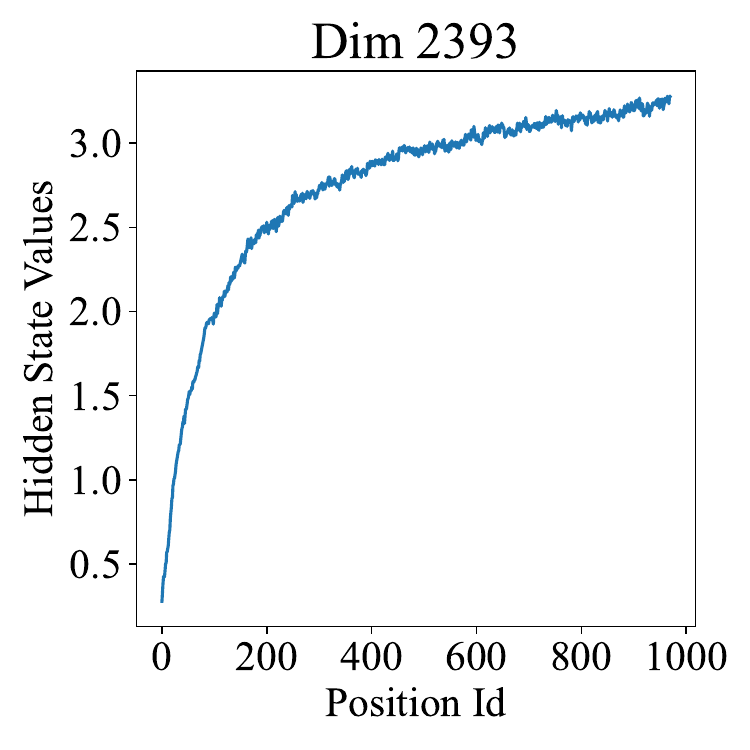}}
  \subfloat[MPT-30b]{
    \includegraphics[height=0.5\columnwidth]{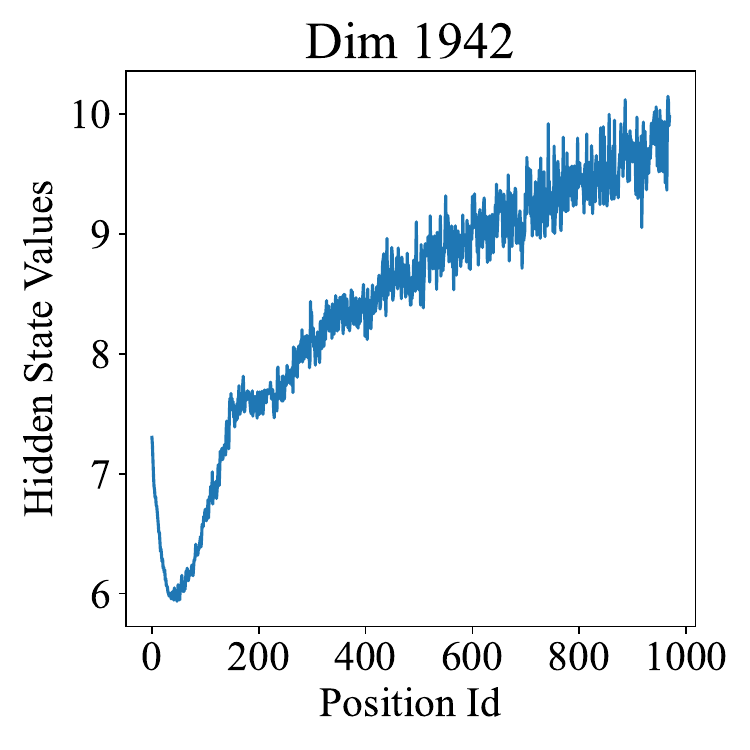}}
  \subfloat[TinyLlama-NoPE-1.1B]{
    \includegraphics[height=0.5\columnwidth]{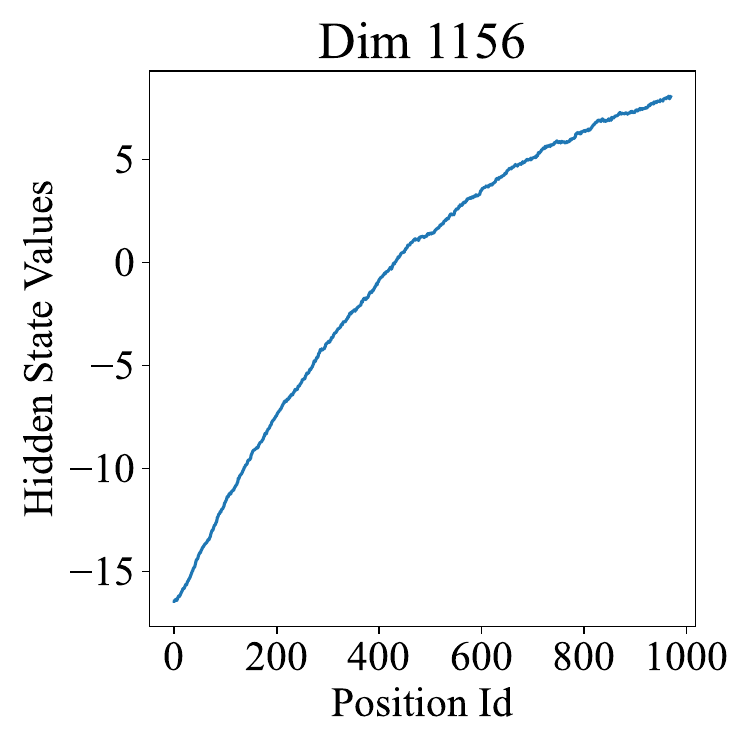}}
  \caption{Hidden states values with the token positions of the positional channel averaged across all layers. 
  }
  \label{fig: Hidden models}
  \vspace{-10pt}
\end{figure*}

\subsection{Position Information can be clearly manifested in Specific Hidden states Channels}

\label{subsec:positional_hidden}
\begin{definition}[Positional Hidden States]\label{def:positional_hidden}
Let $h_k(p)$ denote the $k$-th dimension of the hidden states across each token's position $p$. We define positional hidden states $h_t$ as hidden states whose values change consistently and monotonically with the token position. Therefore, their derivative (after curve fitting) should always be positive or negative:

  \begin{itemize}
     \item $h_t'(p) >0$, $\forall p$ or $h_t'(p) <0$, $\forall p$
 \end{itemize}

 \end{definition}

We have found that the positional information in hidden states can affect position bias, and previous works \cite{haviv_transformer_2022} have proven positional information can be linearly probed from hidden states. However, the high-dimensional hidden states make it hard to concentrate on certain low-dimensional subspace to directly perceive or manipulate positional information in LLMs. To make this easier, we make a strong assumption: the positional information (at least part of it) in the hidden states should be manifested as activation values linearly correlated with each token's absolute position. Thus, on average, there should exist some channels whose values change roughly monotonically with the absolute token positions, as shown in Definition \ref{def:positional_hidden}. We call those channels \textbf{positional channels} and the parts of hidden states belonging to these channels as \textbf{positional hidden states}.

Based on this assumption, to identify positional channels, we average the model's hidden states using 2000 randomly generated strings as inputs, whose lengths are 1000, to make an average hidden states with shape $[sequence\ length, hidden\ size]$. Then we traverse each channel (along $hidden \ size$ axis, in other words, $hidden \ size$ is the number of channels) to check whether its activation values (a one-dimensional array of $sequence\ length$) are changing monotonically with token positions. To eliminate the effect of outliers and noise, we apply sliding window average with window size of 100 to that array and discard the first 30 tokens (because the hidden states values of these tokens are often too huge \citep{sun_massive_2024}) before checking monotonicity.

As the results shown in Figure~\ref{fig: Hidden models}, even though our assumption is very strong, our experiments reveal that causal LMs consistently possess such hidden states across most layers (details in  Appendix~\ref{app:hidden visual}), regardless of whether the model has position embeddings. We further demonstrate that position hidden states are mainly determined by the causal mask but not position embeddings through perturbation experiments in Appendix \ref{app:Perturbation}.

Finally, we aim to confirm that even a single position channel can significantly affect position bias. So we conduct an experiment on KV retrieval, where we manually modify (subtract a fixed value from them) the values in the 213rd hidden states channel of Mistral-7b of a KV pair, and find the attention to this KV indeed greatly increases, which proves our posit. Details about this experiment are in Appendix \ref{app:hidden and attn 26}.

\section{Mitigating Position Bias}

Previous work focusing on PE often attempted to refine the application of RoPE, essentially minimizing the gap in position information (introduced by RoPE) between tokens in different positions.

In Section~\ref{sec:motivation}, we have identified positional information in hidden states as another factor causing position bias, and found positional channels strongly related to positional information. Thus, similarly, to minimize the gap in position information (introduced by positional information in hidden states) between tokens, the most direct way is to scale the activation values of those channels.

Therefore, we propose a method to mitigate position bias by scaling the positional hidden states, as shown in Figure \ref{fig:scale method}. Specifically, it consists of two steps: identifying the positional hidden states $h_{t}$ and scaling (multiplying) them by a factor $s$.

\subsection{Problem Formulation}

Given a pre-trained LLM $\boldsymbol{\theta}$ and a general dataset $\{\bm{x}, \bm{y}\}$, our objective is to find the optimal positional hidden states $h_{t}$, i.e. the $t$-th channel of the hidden states, and the corresponding scaling factor $s$ to maximally reduce position bias, which can be formulated as follows:

\vspace{-5pt}
\begin{equation}
    \begin{aligned}
        \mathop{\arg\min}_{h_t \in \mathcal{H}, s<1} \frac{1}{|\bm{P}|}\sum_{i=1}^{|\bm{P}|} \mathcal{L}\left(\bm{x}, \bm{y}, \bm{p}_i ; F(\boldsymbol{\theta}, h_t, s)\right) \\
    \end{aligned}
    \label{eq:objective}
\end{equation}

where $\bm{P}$ represents the set of different positions $p_i$ of the key information within the context of the prompt $\bm{x}$, $F(\boldsymbol{\theta}, h_t, s)$ denotes the operation of scaling the LLM $\boldsymbol{\theta}$ on the $t$-th channel ($t \in [1,hidden\ size]$) of its hidden states by the scaling factor $s$ (the specific scaling method is in Section \ref{sec:spe method}), and $\mathcal{L}$ denotes the loss for general downstream tasks of the modified model.

In practice, due to the computing cost, the loss is computed on a small validation dataset of some representative tasks. Intuitively, we can traverse all different values of $t$ and obtain the loss on a calibration (validation) dataset to find the optimal one. However, due the large $hidden\ size$ which is typically over 4k, traversal search is too time-consuming, taking nearly 3 days for a 7B LLM. So, we design a search algorithm, mainly based on the monotonicity feature mentioned in Section~\ref{sec:motivation}, to first select a small set (no more than 50) of channels that are most likely to carry positional information that can affect position bias. Then we only need to traverse the small set of candidates to obtain their losses.

\subsection{Identifying Positional Hidden States}

\begin{figure}
\begin{algorithm}[H]
\begin{algorithmic}[1]
\STATE {\bfseries Input:} LLM $\bm{\theta}$, hidden states $\mathcal{H}$, layer number $L$, validation set $\mathcal{D}_{\text{val}}$, positions set $\bm{P}$, threshold $\varepsilon$

\LineComment{Indentify top-K positional dimensions} %
\STATE $\bm{\rho} \gets \phi$
\FOR{$t \gets 1$ to $|\mathcal{H}|$}
\STATE $c_t \gets 0, g_t \gets 0$ 
\FOR{$l \gets 1$ to $L$}
\IF{$h_t'(p) > 0, \forall \ p$ or $h_t'(p) < 0, \forall \ p$} 
\STATE $c_t \gets c_t + 1, g_t \gets  g_t + \text{Smooth}(h_t^l)$\\
\ENDIF
\ENDFOR
\IF{$c_t > \varepsilon$}
\STATE $\bm{\rho} \gets \bm{\rho}\cup \{t\}$\\
\ENDIF
\ENDFOR
\STATE $\bm{\rho} \gets \mathop{\arg\min_K}\limits_{t \in \bm{\rho}} g_t$ \\

\LineComment{Evaluate on the validation dataset}
\FOR{$t \in \bm{\rho}$}
\STATE $\mathcal{L}_t \gets 0$
\FOR{$p \in \bm{P}$}
\STATE $\mathcal{L}_t \gets \mathcal{L}_t + \mathcal{L}(\bm{x}, \bm{y}, p;F(\bm{\theta}, h_t, s))$
\ENDFOR
\ENDFOR
\STATE $\overline{t} \gets \mathop{\arg\min_k}\limits_{t \in \bm{\rho}} \mathcal{L}_t$
\STATE $\mathrm{return}\,\,\,\overline{t}$

\end{algorithmic}

  \caption{
  Positional Hidden State Search
  }
  \label{alg:position_search}
\end{algorithm}
\end{figure}

We have defined positional hidden states in Definition~\ref{def:positional_hidden}, but it is just an ideal situation. In practice, because the hidden states is mainly determined by specific input text, the original values of them will not strictly satisfy monotonicity, which means we cannot rely on strict monotonicity to select the positional channels, but whether it roughly conforms to monotonicity. Thus, we use least square polynomial fit to approximate the values in the channel. Moreover, hidden states of different layers also vary. So we consider this channel as roughly monotonic if the fitted curve conforms to monotonicity in more than a quarter of all the layers of the model.

Using curve fitting, we can usually identify dozens or hundreds of channels that exhibit various degrees of monotonicity. However, some of them are very close to the ideal monotonic curve, while others fluctuate violently although they are roughly monotonic. So we use another indicator, smoothness, to evaluate if how close they are to the ideal monotonic change. The smoothness score is calculated by $\int |h''_t(p)|^2$, where $h''(p)$ is the second derivative (or difference) of $h(p)$, and a smaller score means smoother. Only when $h_t$ is considered roughly monotonic, its smoothness score will be calculated, and averaged across layers. Then we only maintain the channels with the Top-$K$ smoothness score. $K$ is set to 10 in our default setting. 

Finally, we evaluate the average loss across $K$ channels using a 100-sample calibration dataset for KV retrieval. The channel with the lowest loss is chosen for scaling.

To determine the optimal scale factor, we perform a grid search over \{0.5, 0, -0.5, -1\}, selecting the factor with the lowest loss.

We organize our search algorithm in Algorithm~\ref{alg:position_search}, the search process consists of the following two steps: 
1) Identify all the channels $\rho$ of the hidden states that are roughly monotonic in more than $\varepsilon$ layers, and select the top-$K$ channels with the lowest smoothness score. Here $c_t$ is the number of layers where $h_t(p)$ is monotonic, and $g_t$ is the smooth score of $h_t(p)$. 
2) Use a small calibration dataset $\mathcal{D}_{\text{val}}=\{\bm{x},\bm{y}\}$ to evaluate the impact of scaling these positional hidden states respectively and select the channel $h_{\overline{t}}$ that can lead to the minimal loss $\mathcal{L}_{\overline{t}}$.

\subsection{Scaling Positional Hidden States}
\label{sec:spe method}
\begin{figure*}
    \centering
    \includegraphics[width=0.88\linewidth]{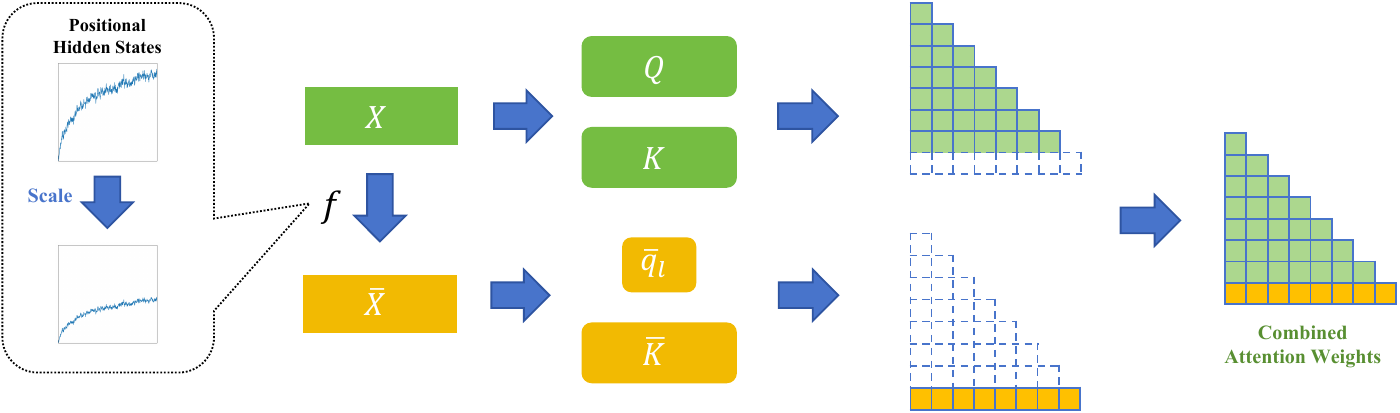}
    \caption{The framework of scaling positional hidden states and modifying attention.}
    \label{fig:scale method}
\end{figure*}

In our early experiments we find simply scaling the positional channel of the hidden states in every layer or in every module of the model can lead to unstable performance of the model.
Therefore, to minimize the side-effect of scaling positional hidden states, our proposed method is finely designed to scale the positional hidden states only affecting the attention weights from last token of the sequence, as shown in Figure~\ref{fig:scale method}. 

Due to the inherent lack of interpretability in the working mechanisms of LLMs, our design is mainly based on the experience of trial and error experiments. So, we use some ablation experiments, shown in Appendix \ref{app:ablation}, to explain why we scale only 1 channel but not more, modify the last token's attention but not all tokens, and use 3 indicators to search the optimal channel.

Specifically, in our method, for the tokens preceding the last token, the attention calculation remains the same as the original. In a sequence of length $l$, for the last token's attention computation, we obtain the modified query state $\bm{\overline{q}}_{l}$ (of the $l$-th token, i.e. the last token) and key states $\bm{\overline{K}}$ (of all the tokens) by scaling the positional hidden states. That is,
\begin{equation}
\begin{split}
    &\bm{\overline{q}}_{l} = \mathcal{P}(W^Qf(\bm{h}(l),p, s), l) \\ 
    &\bm{\overline{K}} = \mathcal{P}(W^Kf(\bm{h},p, s), [1,2,...,l])
\end{split}
\end{equation}

Here $f(\bm{h},p, s)$ means the $p$-th channel of $\bm{h}$ is scaled by the factor $s$. 
Therefore, the combined attention calculation is as follows:
\begin{equation}
    \begin{aligned}
        \bm{z} =
        \left\{
	\begin{aligned}
		&\text{Softmax}(\frac{\bm{q}_{i}\bm{K}^\top + \text{Mask}}{\sqrt{d}})\bm{V} , &  i < l\\
		&\text{Softmax}(\frac{\bm{\overline{q}}_{l}\bm{\overline{K}}^\top }{\sqrt{d}})\bm{V} , & i = l
	\end{aligned}
	\right.
    \end{aligned}
    \label{eq:scale_factor}
\end{equation}
where ${\bm{z}}$ is the attention output.

Except calculating the combined attention weights, the other modules remain the same as those in the original method. We implement our method using FlashAttention-2~\citep{dao2023flashattention} with minimal overhead, so our approach results in only a slight increase in latency, as shown in Appendix \ref{app:time}, 

Moreover, we find modifying the hidden states in the initial layers (especially the 1st or 2nd layer) will also make the model's output more unstable, as well as the last layers. Thus we only apply our scaling to layers in the middle (they are more likely to be retrieval-related layers), for example, the 10th to 25th layers with total 32 layers. The specific layer selection is also based on engineering experience.

\section{Experiments}

\subsection{Setup}
\label{subsec:setup}

\paragraph{Evaluation Tasks and Models}
We apply our method to a wide range of popular open-source LLMs, including:
1) RoPE~\citep{chen_extending_2023} models: LLaMA-2 (7B, 13B)~\citep{touvron2023llama}, Mistral-7B~\citep{jiang2023mistral}, Gemma-7B~\citep{team2024gemma}, Qwen1.5-7B~\citep{qwen};
2) Context window extended models: Vicuna-16k (7B, 13B)~\citep{vicuna2023};
3) Alibi~\citep{press_train_2022} models: MPT-30B~\citep{MosaicML2023Introducing}. All the models are instruction-tuned versions.

And we evaluate the performance in 2 aspects:
1) Position-bias-related tests: ``lost in the middle'' benchmark, including NaturalQuestion multi-document QA~\citep{liu_lost_2023} and KV retrieval~\citep{liu_lost_2023} with the Gold document or KV at different positions in the context. The multi-document QA task includes 20 documents with a prompt length of about 2.3k tokens, while the KV retrieval task includes 140 KV pairs with an average length of about 10k tokens.
2) General long-context multi-task benchmark: LongBench~\citep{bai2023longbench}, including multi-document QA, single-document QA, summarization, few-shot learning, synthetic tasks, and code completion, totaling 16 tasks with an average length of 37k tokens.

For prompts that exceed the context windows of LLMs, we follow LongBench's approach by truncating from the middle and retaining the head and tail of the prompt to fit within the context windows.
We use the metrics and scripts provided by the benchmark's github repository for evaluation. More details about the benchmarks are in Appendix \ref{app:dataset detail}.

\paragraph{Implementation Details} We implement our approach using PyTorch and HuggingFace Transformers in an A100 GPU. To ensure stable and reproducible results, we use greedy decoding in all generation experiments.

In the searching algorithm, we set the top-$K$ to top-10 and $\varepsilon$ to $L/4$, where $L$ is the number of total layers. The search process takes approximately 10 minutes. 
The details of the scaling channels, layers, and factors are shown in Appendix~\ref{app:scale dim}.

\paragraph{Baselines}  
Besides the original models, we include another training-free positional bias mitigation method as the baseline, which is signified by \textbf{w/ Ms-PoE}~\citep{zhang_found_2024}. It is a head-wise position embedding scaling method, and we follow its default settings to apply scaling coefficients of 1.2 to 1.8 starting from the 3rd layer.

\subsection{Main Results}
\begin{table*}[tb]
    \small
    \centering
    \setlength{\tabcolsep}{1mm}
    \vspace{-2ex}

    \resizebox{1.9\columnwidth}{!}{
    \begin{tabular}{l|cccccc|cccccc}
    \toprule
        \multirow{2}{*}{Methods} &  \multicolumn{6}{@{}c}{{\bf NaturalQuestion}} &  \multicolumn{6}{@{}c}{{\bf KV Retrieval}} \\
        & 1st & 5th & 10th & 15th & 20th & Avg. & 0\% & 25\% & 50\% & 75\% & 100\% & Avg. \\
    \midrule
    LLaMA-2-7b-chat &  32.4&23.8&30.6&31.6&38.2&31.3&77.6&24.6&62.0&35.6&78.0&55.6\\
    LLaMA-2-7b-chat w/ Ms-PoE & \textbf{40.8}&29.2&33.0&32.8&39.6&35.1&\textbf{95.0}&29.8&21.4&\textbf{51.8}&89.8&57.6\\
    LLaMA-2-7b-chat w/ Ours &33.6 & \textbf{34.0} & \textbf{40.6}& \textbf{43.0} & \textbf{51.8}& \textbf{40.6} &63.6&\textbf{38.0}&\textbf{82.2}&40.6&\textbf{94.6}&\textbf{63.8}\\
    \midrule
    LLaMA-2-13b-chat & 45.2&39.6&40.4&44.2&51.0&44.1  &74.2 &\textbf{39.0}&\textbf{70.4}&\textbf{84.4}&\textbf{86.8}&\textbf{71.0}\\
    LLaMA-2-13b-chat w/ Ms-PoE &48.4&41.4&42.4&45.4&52.6&46.0&\textbf{87.8}&28.0&35.4&49.2&83.0&56.7 \\
    LLaMA-2-13b-chat w/ Ours & \textbf{50.6}&\textbf{43.4}&\textbf{45.0}&\textbf{49.4}&\textbf{58.2}&\textbf{49.3}&41.2&17.0&49.6&76.8&84.8&53.9\\
    \midrule
    Vicuna-7b-v1.5-16k & \textbf{70.4}&54.8&46.8&45.8&47.8&53.1&\textbf{98.4}&0.8&0.2&0.2&0.2&20.0\\
    Vicuna-7b-v1.5-16k w/ Ms-PoE &67.0&55.2&50.6&46.8&48.2&53.6&97.4&\textbf{36.8}&\textbf{15.6}&\textbf{5.2}&6.6&\textbf{32.3} \\
    Vicuna-7b-v1.5-16k w/ Ours &63.8&\textbf{57.6}&\textbf{53.6}&\textbf{51.2}&\textbf{55.6}&\textbf{56.4}&95.4&22.0&12.6&\textbf{5.2}&\textbf{20.4}&31.1 \\
    \midrule
    Vicuna-13b-v1.5-16k & 67.4&48.2&45.2&45.6&44.4&50.2&95.6&74.2&64.2&58.8&18.2&62.2\\
    Vicuna-13b-v1.5-16k w/ Ms-PoE &\textbf{70.0}&\textbf{51.4}&46.8&42.8&47.0&51.6&91.8 & 59.4 & 71.6 & \textbf{74.4} & \textbf{48.8} & 69.2\\
    Vicuna-13b-v1.5-16k w/ Ours &67.4&\textbf{51.4}&\textbf{47.6}&\textbf{48.8}&\textbf{48.0}&  \textbf{52.7} &\textbf{97.2}&\textbf{83.4}&\textbf{80.8}&68.8&35.4&\textbf{73.1} \\
    \midrule
    Mistral-7b-Instruct-v0.2 &57.2&55.0&61.2&\textbf{61.6}&62.6&59.5 & \textbf{99.8}&93.0&89.0&95.0&94.2&94.2\\
    Mistral-7b-Instruct-v0.2 w/ Ms-PoE &58.2 & \textbf{60.0} & 62.6 & 58.8 & 62.2 & 60.4 & \textbf{99.8} & \textbf{95.6} & 88.4 & \textbf{96.0} & \textbf{95.4} & \textbf{95.0}\\
    Mistral-7b-Instruct-v0.2 w/ Ours &\textbf{61.2}&56.4&\textbf{63.2}&59.8&\textbf{64.0}&\textbf{60.9}&97.6&93.2&\textbf{90.6}&95.6&93.8&94.2\\
    \midrule
    Gemma-1.1-7b-it &29.6&25.2&28.2&29.6&27.4&28.0&\textbf{98.6}&67.0&62.4&83.4&\textbf{100.0}&82.3\\
    Gemma-1.1-7b-it w/ Ms-PoE & 33.8&29.0&31.6&28.6&28.6&30.3&0.0&0.0&0.0&0.0&0.0&0.0\\
    Gemma-1.1-7b-it w/ Ours &\textbf{35.4}&\textbf{31.4}&\textbf{36.0}&\textbf{35.4}&\textbf{35.0}&\textbf{34.6}&97.6&\textbf{95.8}&\textbf{97.6}&\textbf{96.8}&99.6&\textbf{97.5}\\
    \midrule
    Qwen1.5-7b-chat & \textbf{72.4}&53.8&52.2&51.2&54.4&56.8&\textbf{100.0}&\textbf{97.2}&84.6&60.0&56.4&79.6\\
    Qwen1.5-7b-chat w/ Ms-PoE &67.4&49.8&48.2&47.4&47.0&52.0& 3.4&1.4&2.8&2.6&0.6&2.2\\
    Qwen1.5-7b-chat w/ Ours & 67.4&\textbf{55.2}&\textbf{53.6}&\textbf{56.0}&\textbf{59.4}&\textbf{58.3}&97.2&95.6&\textbf{98.8}&\textbf{76.6}&\textbf{94.4}&\textbf{92.5}\\
    \midrule
    MPT-30b-chat & \textbf{75.6}&\textbf{49.6}&39.0&33.4&39.6&47.4&71.4&34.8&31.6&41.6&74.0&50.7\\
    MPT-30b-chat w/ Ms-PoE & / &/ &/ &/ &/ &/ &/ &/ &/ &/ &/ &/\\
    MPT-30b-chat w/ Ours & 75.0&48.8&\textbf{41.6}&\textbf{40.6}&\textbf{44.0}&\textbf{50.0}&\textbf{99.0}&\textbf{65.8}&\textbf{48.6}&\textbf{46.6}&\textbf{69.4}&\textbf{65.9}\\
    \bottomrule
    \end{tabular}
    }
    \caption{Performance of different methods with different models on NaturalQuestions (20 docs)~\citep{liu_lost_2023} and KV retrieval (140 KV pairs)~\citep{liu_lost_2023}  dataset. ``/ '' signifies ``not applicable''.
    }
        \label{tab:main_results_qa}
\end{table*}

\begin{table*}[htb]
    \small
    \centering
    \setlength{\tabcolsep}{1mm}
    \vspace{-2ex}

    \resizebox{1.8\columnwidth}{!}{
    \begin{tabular}{l|ccccccc}
    \toprule
        Models & SingleDoc & MultiDoc & Synth. & Summ. & FewShot &  Code & AVG \\
         \cmidrule (r){1-1}\cmidrule (lr){2-8}
        LLaMA-2-7b-chat & 28.9 &29.7 & 6.6 & 26.3 & 61.2 & 47.1 &33.3\\
        LLaMA-2-7b-chat w/ Ms-PoE & \textbf{29.8} & \textbf{31.7} & \textbf{10.5} & \textbf{26.7} & 61.0 & \textbf{48.1} & \textbf{34.6}\\
         LLaMA-2-7b-chat w/ Ours & 29.2 & 29.3 & 9.7 & 25.0 & \textbf{61.6} & 46.9 & 33.6 \\
        \cmidrule (r){1-1}\cmidrule (lr){2-8}
        LLaMA-2-13b-chat & 21.4 & 14.6 & 11.2 & 26.1 & 61.5 & 39.8 & 29.1\\
        LLaMA-2-13b-chat w/ Ms-PoE & 20.8 & \textbf{15.4} & \textbf{12.7} & \textbf{27.3} & \textbf{62.8} & 36.3 & 29.2\\
        LLaMA-2-13b-chat w/ Ours & \textbf{30.6} & 9.6 &10.8 & 25.7 & 62.6 & \textbf{43.2} & \textbf{30.4}\\
        \cmidrule (r){1-1}\cmidrule (lr){2-8}
        Vicuna-7b-v1.5-16k & 30.2 & 21.6 & 7.2 & 26.7 & 53.9 & 40.5 & 30.0\\
        Vicuna-7b-v1.5-16k w/ Ms-PoE & \textbf{32.3} & \textbf{24.2} & 8.3 & \textbf{28.0} & \textbf{55.2} & \textbf{43.1} & \textbf{31.8} \\
        Vicuna-7b-v1.5-16k w/ Ours & 27.1 & 22.1 & \textbf{11.2} & 26.1 & 55.0 & 40.2 & 30.3\\
        \cmidrule (r){1-1}\cmidrule (lr){2-8}
        Vicuna-13b-v1.5-16k & 31.1 & 33.8 & 21.2 & 26.2 & 62.0 & 39.8 & 35.7\\
        Vicuna-13b-v1.5-16k w/ Ms-PoE & \textbf{34.5} & 33.1 & 16.0 & \textbf{27.5} & \textbf{64.5} & 37.6 & 35.5\\
        Vicuna-13b-v1.5-16k w/ Ours & 30.1 & \textbf{35.1} & \textbf{25.0} & 25.8 & 63.5 & \textbf{41.7} & \textbf{36.9} \\
        \cmidrule (r){1-1}\cmidrule (lr){2-8}
        Mistral-7b-Instruct-v0.2 & 37.8 & 28.5 & 49.7 & 28.8 & \textbf{65.3} & \textbf{52.9} & 43.8\\
        Mistral-7b-Instruct-v0.2 w/ Ms-PoE & \textbf{41.7} & 22.2 & 38.4 & 2.8 & 23.8 & 19.5 & 24.7\\
        Mistral-7b-Instruct-v0.2 w/ Ours & 38.4 & \textbf{30.4} & \textbf{49.8} & \textbf{29.4} & 64.8 & \textbf{52.9} & \textbf{44.3}\\
        \cmidrule (r){1-1}\cmidrule (lr){2-8}
        Gemma-1.1-7b-it & 39.4 & \textbf{23.2} & 32.2 & 24.2 & 14.4 & \textbf{19.8} & 25.5\\
        Gemma-1.1-7b-it w/ Ms-PoE & \textbf{41.7} & 22.2 & \textbf{38.4} & \textbf{24.9} & 14.0 & 19.5 & \textbf{26.8}\\
        Gemma-1.1-7b-it w/ Ours & 39.0 & 23.0 & 35.5 & 24.5 & \textbf{14.9} & 19.3 & 25.7\\
       \cmidrule (r){1-1}\cmidrule (lr){2-8}
        Qwen1.5-7b-chat & \textbf{46.4} & 39.5 & 38.4 & 22.3 & 56.4 & \textbf{50.2} & \textbf{42.2}\\
        Qwen1.5-7b-chat w/ Ms-PoE & 42.0 & \textbf{41.5} & 30.3 & \textbf{25.7} & 46.5 & 38.0 & 37.3 \\
        Qwen1.5-7b-chat w/ Ours & 45.8 & 38.8 & \textbf{38.5} & 22.1 & \textbf{57.6} & 49.6 & \textbf{42.2}\\
        \cmidrule (r){1-1}\cmidrule (lr){2-8}
        MPT-30b-chat & 27.9 & \textbf{21.9} & \textbf{7.5} & 25.7 & 57.3 & 39.3 & \textbf{29.9}\\
        MPT-30b-chat w/ Ms-PoE & /&/&/&/&/&/&/\\
        MPT-30b-chat w/ Ours & \textbf{29.4} & 19.5 & 6.7 & \textbf{25.8} & \textbf{57.6} & \textbf{40.1} & \textbf{29.9}\\
    \bottomrule
    \end{tabular}
    }    \caption{Performance of different methods with different models on LongBench~\citep{bai2023longbench}.}
    \label{tab:long_bench}
\end{table*}

From the evaluation results of ``lost in the middle'' benchmark in Table~\ref{tab:main_results_qa}, several conclusions can be drawn:
1) Our method better improves overall performance at most positions, with the average improvement of up to 9.3\%, 15.2\% in NQ and KV retrieval, respectively, except for LLaMA-2-13B in KV retrieval. 
2) Our method effectively enhances LLMs' exploitation of information located in the middle and rear parts of the prompt. When key information is at the beginning of the prompt, performance is sometimes increased or decreased. Considering only the average performance of the last four positions, our method's improvements over the original increase to 11.3\% and 16.8\% in NQ and KV retrieval, respectively, much higher than MsPoE.
3) Our method demonstrates better generalization performance, showing improvement on nearly all types of models, regardless of RoPE models, context window extended models like Vicuna-16K, or Alibi models like MPT. In contrast, Ms-PoE causes instability of the output of Qwen and Gemma in KV retrieval, leading to very low accuracy.

From the evaluation results of Longbench in Table~\ref{tab:long_bench}, our method demonstrates varying degrees of improvement across different tasks, with the most significant increases being 1.5\% in few-shot learning tasks, 3.4\% in code tasks, 4\% in synthetic tasks, 9.2\% in single document QA tasks, and 1.9\% in multi-document QA tasks. However, maybe because Longbench mainly focuses on comprehensiveness and reality, the influence of position bias on these tasks is relatively minimal, our method does not significantly improve the average scores (only a little). But it at least demonstrates that it will not impair the model's original capability to handle various long context tasks.

\subsection{Analysis}
\paragraph{From Bias to Balance}
As shown in Table \ref{tab:main_results_qa}, there is an trend that our method generally increases performance when the key information is at the middle or end, but decreases when at the beginning. It reveals a possible fact that the positional hidden states may be an important factor causing the model to overly focus on the beginning parts of the context while miss the rear or middle parts. Therefore, scaling positional hidden states by a scaling factor less than 1 can reduce its impact, thus shift the model's attention distribution from the bias to the beginning to a more balanced distribution.

\begin{figure}[htb]
    \centering
    \subfloat[w/ Attention]{
    \includegraphics[width=0.5\linewidth]{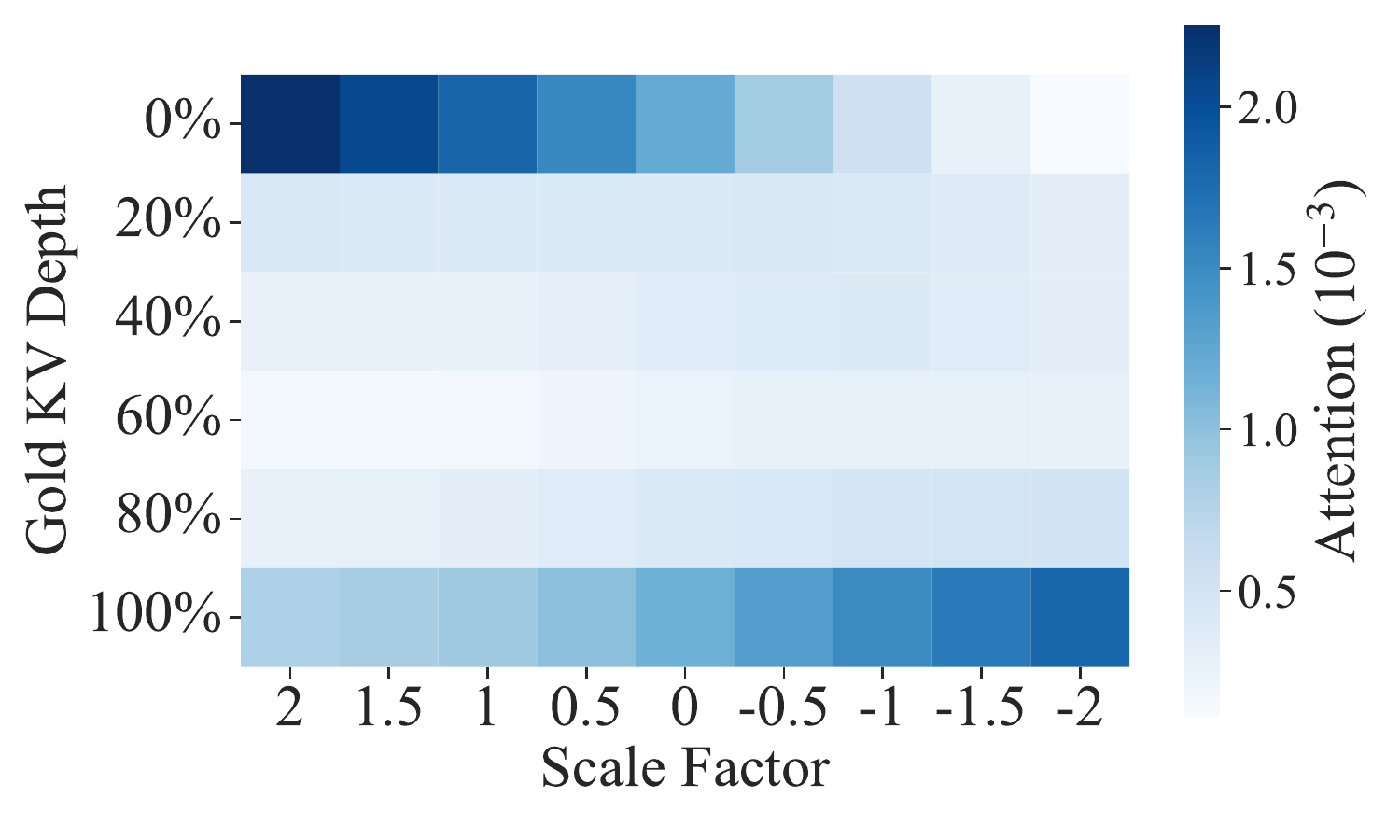}
    \label{sfig: scale factor 1}}
    \subfloat[w/ Performance]{
    \includegraphics[width=0.49\linewidth]{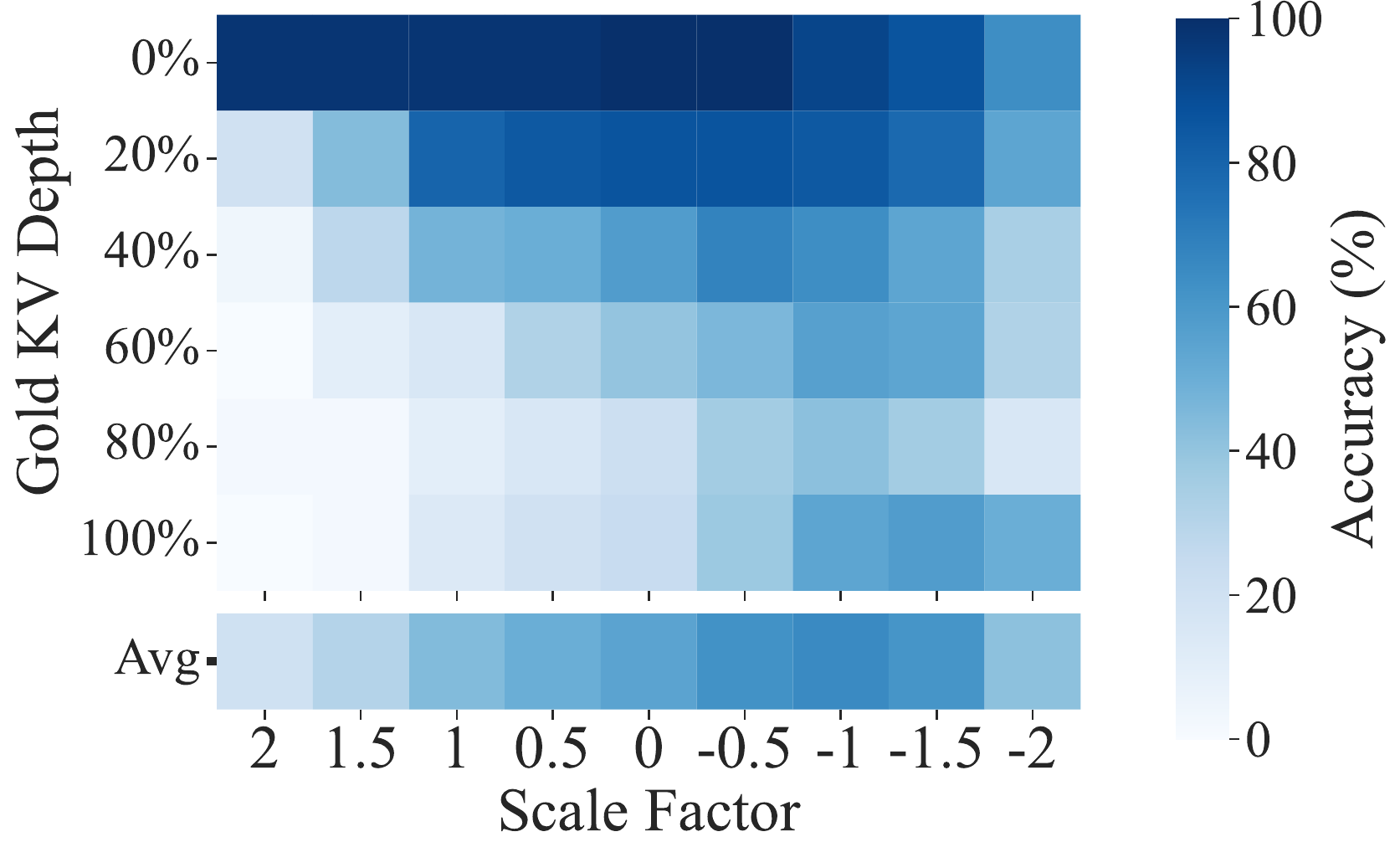}
    \label{sfig: scale factor 2}}
    \caption{Attention distribution on KVs at different positions (depths) and performance when scaling the 2,393rd channel of Vicuna-7b with different scale factors on KV retrieval~\citep{liu_lost_2023} of 100 KV pairs.}
    \label{fig:scale factor}
\end{figure}

In theory, a scaling factor over 1 amplifies the effect of positional hidden states, and one between 0 and 1 shrinks it. A negative factor can even reverse it. Correspondingly, as shown in Figure~\ref{fig:scale factor}, a positive scaling factor over 1 causes the model to focus more on the beginning, while a negative factor shifts the focus towards the end. A factor between 0.5 and -1 leads to a relatively balanced attention distribution, where the average accuracy also peaks. These results demonstrate that scaling positional hidden states are essentially steering the position information in the model, to shift the position bias towards our desired direction.

\begin{table}
	\centering

  \resizebox{\columnwidth}{!}{
	\begin{tabular}{l|cc}
		\toprule
		Model & MMLU &Reorder\\
            \midrule
	Vicuna-7b-v1.5-16k &48.22 &20.83 \\
    Vicuna-7b-v1.5-16k w/ Ours & 48.38&20.83  \\
        \midrule
    Qwen1.5-7b-chat &60.84 &28.13  \\
    Qwen1.5-7b-chat w/ Ours & 61.43&28.13  \\
        \midrule
    Mistral-7B-Instruct-v0.2 &60.31 &18.75  \\
    Mistral-7B-Instruct-v0.2 w/ Ours &60.38 &19.79  \\
            \bottomrule
	\end{tabular}
 }
  	\caption{Performance of difference models on MMLU and the timeline reorder task, before and after applying our method.}
\label{tab:timeline order}
\end{table}

\paragraph{Side Effects}

It is natural to concern scaling hidden states may impair the model's normal workflow.
So we utilized MMLU benchmark \citep{hendrycks_measuring_2021}, which assesses LLMs' general capabilities, and the timeline-reorder task from LooGLE \citep{li_loogle_2023}, which requires arranging events chronologically across extensive text and is sensitive to positional information, to evaluate whether our method will impair the original abilities of LLMs. As shown in Table \ref{tab:timeline order}, there is no significant detriment to the models' performance with our methods in default settings.

\section{Conclusion}

We propose a method which scales positional hidden states to mitigate the position bias issue of LLMs. Specifically, our study first explore the relationships between position bias (in the model's performance), attention weights, causal mask, positional information in hidden states and positional channels, to confirm the positional channels can affect position bias.
Based on these findings, we design a positional channel search algorithm to identify positional hidden states, and mitigate the model's position bias by scaling the positional hidden states. These findings provide a new perspective for research related to position information and position bias of LLMs.

\section*{Limitations}
The selection of layers to apply is mainly based on engineering experience. And although our method has tried to minimize the side effect of scaling hidden states, if choosing a too large scale factor or an inappropriate channel to scale, we still observe significant performance degradation.

Some newly released models like Qwen2.5 \cite{team_qwen25_2024} have achieved nearly perfect performance on retrieval-based long-context tasks. Thus, on traditional benchmarks, our method may not bring more improvement. Experiment on more newly released models and benchmarks may still be needed in the future.

\bibliography{ref}

\appendix
\section{Related Works}
\label{sec:related}

\paragraph{Addressing Position Bias}
Recent works reveal that LLMs often exhibit position bias, also known as the "lost in the middle" phenomenon~\citep{liu_lost_2023,kamradt2023needle}. Previous efforts to mitigate this bias fall into several categories: 
1) RoPE-based methods: These approaches modify the RoPE computation process to alleviate long-distance attention decay (less attention means less information retrieved), including Attention Bucket~\citep{chen_fortify_2023}, which uses an ensemble of multiple RoPE bases to mitigate position bias, and Ms-PoE~\citep{zhang_found_2024}, which dynamically interpolates with a small coefficient for different heads.
2) SFT-based methods~\citep{junqing2023never,yu_training_2024,an_make_2024}: These methods construct data with more diverse key information distributions or employ system2think SFT tasks to mitigate position bias. They require further training of the model.
3) Attention mask-based methods~\citep{he2024position}: These methods modify attention mechanisms, including
Attention Transition~\citep{gao_empower_2023}, which redirects attention to significant parts of the context and Stable Mask~\citep{yin_stablemask_2024}, which introduces pseudo attention into the causal mask, ensuring stable attention distribution when facing lengthy texts.
4) Prompt-based methods~\citep{jiang_longllmlingua_2023, peysakhovich_attention_2023}: These methods introduce an external module to reorder or compress information in the prompt, thereby mitigating position bias.

\section{Datasets Details}
\label{app:dataset detail}

We choose NaturalQuestion Multi-document QA and Key-Value Retrieval datasets used in ``lost in the middle'' paper \citep{liu_lost_2023} to evaluate the degree to which our method alleviates position bias. NaturalQuestion Multi-document QA requires the model to answer the question based on one key information document which is inserted in a long context consisting of many irrelevant documents. And Key-Value Retrieval needs the model to retrieve the value corresponding to the given key from a list consisting of hundreds of Key-Value pairs. These two datasets are both classic long-context tasks which aim to evaluate the differences of model performance when key information is located at different positions in the context. The evaluation metric is accuracy, based on whether the model's response contains the string of the correct answer.

In addition, we evaluate our method's improvements across multi task types, using LongBench \citep{bai2023longbench}, a benchmark for bilingual, multitask, and comprehensive assessment of long context understanding capabilities of LLMs. It contains six major categories, covering single-document QA, multi-document QA, summarization, few-shot learning, synthetic tasks and code completion. The evaluation metrics are: F1 for single-document QA and multi-document QA, Rouge-L for summarization, accuracy (exact match) for few-shot learning and synthetic tasks, and edit similarity for code completion. During inference, since the original context may sometimes be too long, the input sequences will be truncated in the middle part to avoid exceeding the context window of the model.

\section{Other Implementation Details}
\label{subsec:implemention}
\subsection{Curve Fitting}
When we perform curve fitting on $h(p)$, we use least-squares cubic polynomial fit. And when judging its monotonicity, we skip the first 30 positions because the first a few values are often outliers. Since $h(p)$ is originally a discrete function, in practice, we employ the second-order difference to approximate the second-order derivative when computing smoothness.
\subsection{Ms-PoE on Mistral}
When applying Ms-PoE~\citep{zhang_found_2024} to mistral-7b~\citep{jiang2023mistral} with its default parameters (minimal scale factor is 1.2 and maximal is 1.8), we found the model fail to generate normal responses, so we set the maximal scale factor to 1.2, under which Ms-PoE~\citep{zhang_found_2024} is equal to PI~\citep{chen_extending_2023} with scale factor 1.2.

\subsection{Parameters of the Scaling Method}
\label{app:scale dim}
\begin{table}[htb]
	\centering
  \resizebox{\columnwidth}{!}{
	\begin{tabular}{l|ccc}
		\toprule
		Model &  Channel Index & Scale factor &	Applied layers \\
            \midrule
		LLaMA-2-7b-chat & 2,393 & -1 & 10\textasciitilde25 \\
		LLaMA-2-13b-chat & 4,283 & -1 & 10\textasciitilde34 \\
		Vicuna-7b-v1.5-16k & 2,393 & 0 & 10\textasciitilde25\\
            Vicuna-13b-v1.5-16k  & 4,923 & 0 & 10\textasciitilde34 \\
            Mistral-7B-Instruct-v0.2  & 213 & 0 & 10\textasciitilde25 \\
		Gemma-1.1-7b-it & 1,665 & 0 & 10\textasciitilde22 \\
		Qwen1.5-7b-chat & 1,081 & 0.2 & 10\textasciitilde25 \\
            MPT-30b-chat & 6,926 & 0 & 10\textasciitilde42  \\
		\bottomrule
	\end{tabular}
 }
  	\caption{The scaled channels, scale factors and applied layers of models.}
	\label{tab:setting}
\end{table}

The scaled channel indices, scale factors and applied layers of each model we use in out experiments are shown in Table \ref{tab:setting}.

\subsection{Inference Latency}
\label{app:time}
\begin{table}[htb]
	\centering

	\begin{tabular}{l|cc}
  		\toprule
		Method &  KV Retrieval  & Multi-Doc  \\
            \midrule
		FlashAttention-2 &  22 &  14  \\
            Ours &  32 &  15  \\
		Ms-PoE  & 61 &  26  \\
            \bottomrule
	\end{tabular}
    \caption{Average inference time (minutes) of LLaMA-2-7b-chat in a single A100 on KV retrieval.}
	\label{tab:time consumed}
\end{table}
Table \ref{tab:time consumed} shows the running time of LLaMA-2-7b-chat with different methods in the KV retrieval dataset consisting of 500 samples with average length of about 10,000, and the multi-document QA dataset consisting of 500 samples with average length of about 3,300. Our method requires recompute the query and key states, thus inevitably requires more time compared to baseline, but the cost is within an acceptable range. In contrast, Ms-PoE \citep{zhang_found_2024} need to compute the attention weights twice, resulting in a doubling of time consumption.

\section{Details of The Confirmatory Experiments}

\subsection{Obtain Attention to Key Information}
\label{app:obtain attn}
To avoid the influence of internal knowledge in the model and make attention calculation simpler, we conduct a KV retrieval task, whose prompt format is as follows: 

\begin{tcolorbox}[]		
 Json data: \\
 \{"os08jbk1limft6wgxeda": "imx6lyp4b8ogjaq7ret1", \\
 ......($n$ key-value pairs)\} \\
 \\ 
 The value of key "os08jbk1limft6wgxeda" is "
 \label{kv prompt}
\end{tcolorbox}

The last token of the prompt will directly account for predicting the answer, i.e., the value which need to be retrieved. Hence, the last token's attention weights to the previous text can reflect whether it accurately retrieves the key information. We define the model's attention (in some layer) to the key information as $A_G$ in Eq \ref{avg attention}, where $G$ represents the set of token positions corresponding to where the key information is at, $l$ is the position of the last token of the prompt, and $a_{l,j}$ represents the attention weight of the $l$-th token to the $j$-th token. By shifting $G$, we use the same method to calculate its attention to each other KV pairs.
\begin{equation}
    A_G= \frac{1}{|G|}\sum_{j\in G} a_{l,j}
    \label{avg attention}
\end{equation}

\subsection{Attention is Related to Performance}
\label{app:changing_attention}
\begin{figure*}[htb]
  \centering
  \subfloat[Attention]{
    \includegraphics[height=0.5\columnwidth]{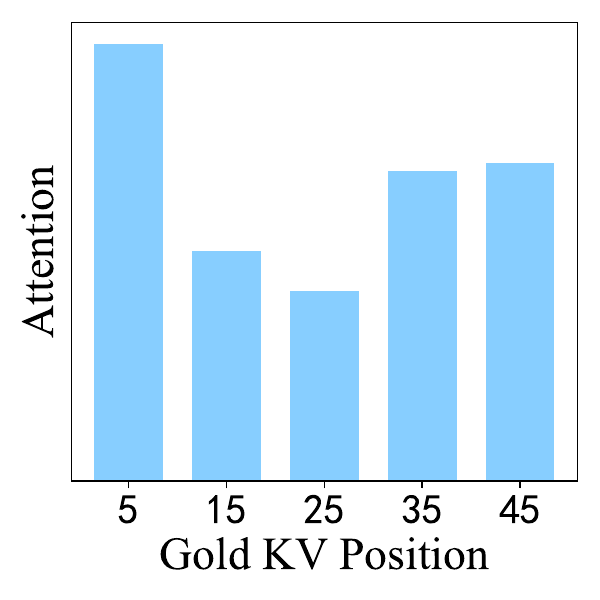}}
  \subfloat[Attention w/ modification]{\label{sfig:attn * 2}
\includegraphics[height=0.5\columnwidth]{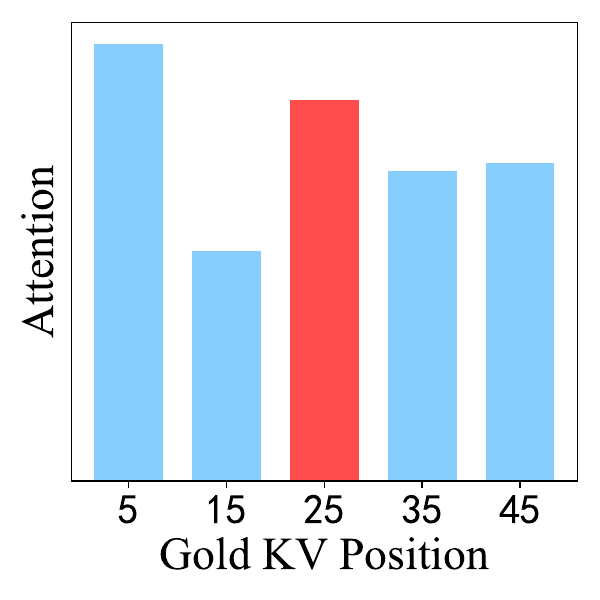}}
  \subfloat[Accuracy]{
    \includegraphics[height=0.5\columnwidth]{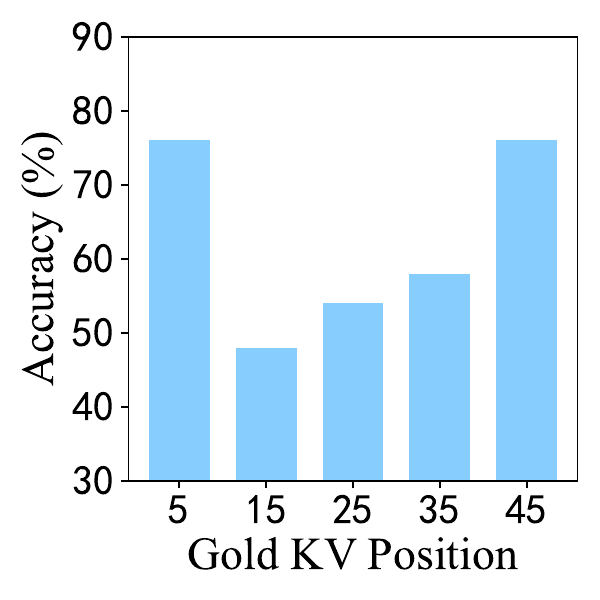}}
  \subfloat[Accuracy w/ modification]{\label{sfig:acc * 2}
    \includegraphics[height=0.5\columnwidth]{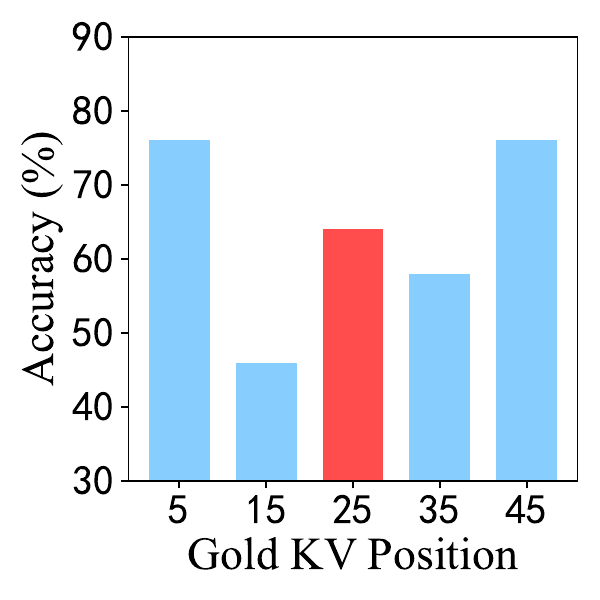}}
  \caption{Distribution of attention weight and accuracy as the gold KV is placed at different positions in the prompt. (b) and (d) are situations when the attention on the 25th KV pair is manually modified.}
  \label{fig: changing_attention}
\end{figure*}

As shown in Figure~\ref{fig: changing_attention}, when we manually double the attention weights of the gold KV (i.e. the 25th KV, as illustrated in Figure \ref{sfig:attn * 2}) during the model's forward pass on the KV retrieval task, the retrieval accuracy when the gold KV is the 25th KV improves, (shown in Figure \ref{sfig:acc * 2}). This demonstrates that the attention weights to the key information are positively correlated with retrieval accuracy.

\subsection{How We Modify Causal Mask and Position Embedding in KV Retrieval}
\label{app:Modify causal mask and PE}
\begin{figure*}[htb]
  \centering
    \includegraphics[height=0.6\columnwidth]{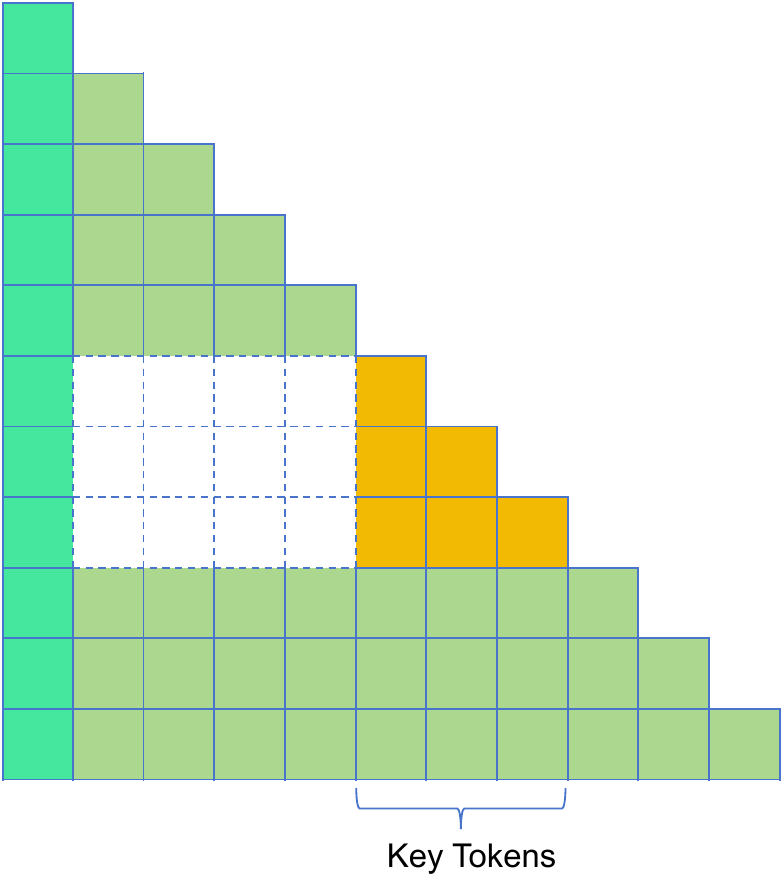}
      \caption{Cropping the causal mask to let key tokens unable to see previous tokens, except the first token.}
        \label{fig: change mask}
\end{figure*}

In the method 1 in section \ref{sec:crop mask}, we crop the causal mask to let the ``key tokens'' unable to attend the previous tokens. As shown in Figure \ref{fig: change mask}, the white part represents the cropped part, which means attention weights are 0, and the orange part represents the attention between tokens within key tokens. In addition, we have retained the attention of key tokens to the first token to maintain the stability of attention distribution. What is more, we only modify the causal mask in layers 1\textasciitilde8, but as the results, the attention to the key information is still significantly improved in layers 15\textasciitilde31, which indicates the positional information generated by causal mask in former layers can be transmitted to latter layers using posisional hidden states as the medium, thus modifying the causal mask solely in the former layers can induce a profound shift in the model's comprehension of positional information.

\begin{figure*}[htb]
  \centering
\includegraphics[height=0.8\columnwidth]{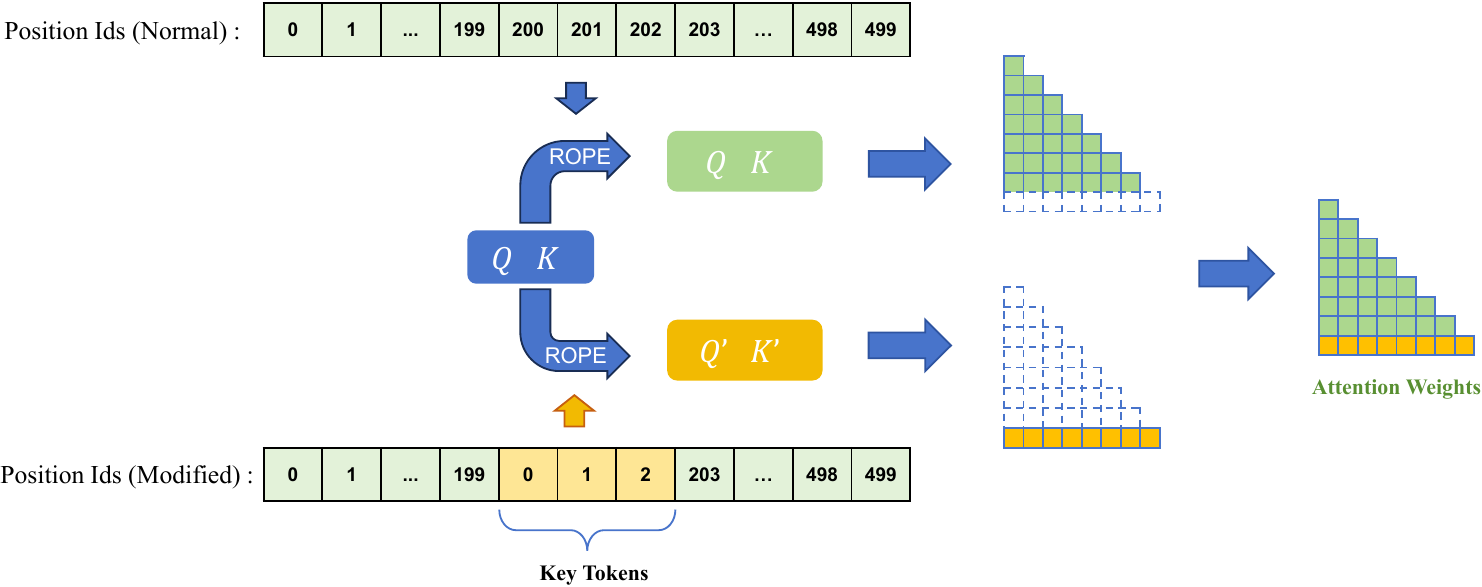}
  \caption{Shifting position ids to the start (PE to beginning).}
    \label{fig: change PE}
\end{figure*}

In the method 2 and 3 in section \ref{sec:crop mask}, we modify the position embeddings through altering the position ids. The specific operation is shown in the Figure \ref{fig: change PE}, in which we directly replace the position ids corresponding to the key tokens with the position ids of the starting tokens (or the ending tokens) , and actually only the attention weights of the last token to previous tokens are modified. We apply this modification in all the layers. Compared to modifying the causal mask, if only modify position embedding in former layers, the attention in the latter layers remains almost unchanged, which indicates the positional information generated by position embedding may be temporary and can hardly be transmitted across layers.

\begin{figure*}[htb]
  \centering
  \subfloat[Hidden States]{
    \includegraphics[height=0.7\columnwidth]{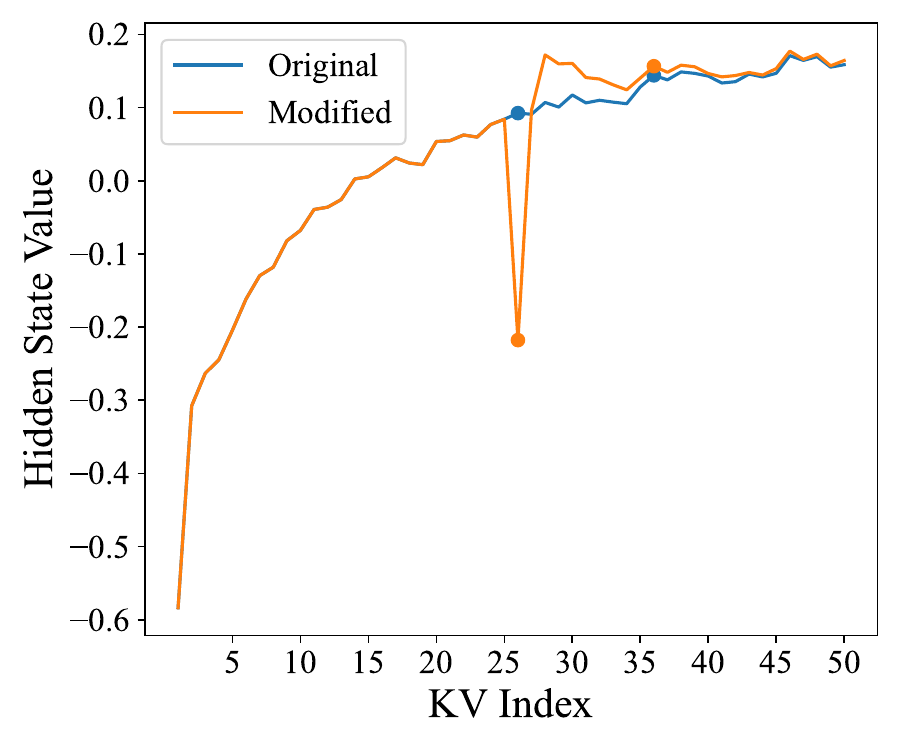}}
  \subfloat[Attention]{
\includegraphics[height=0.7\columnwidth]{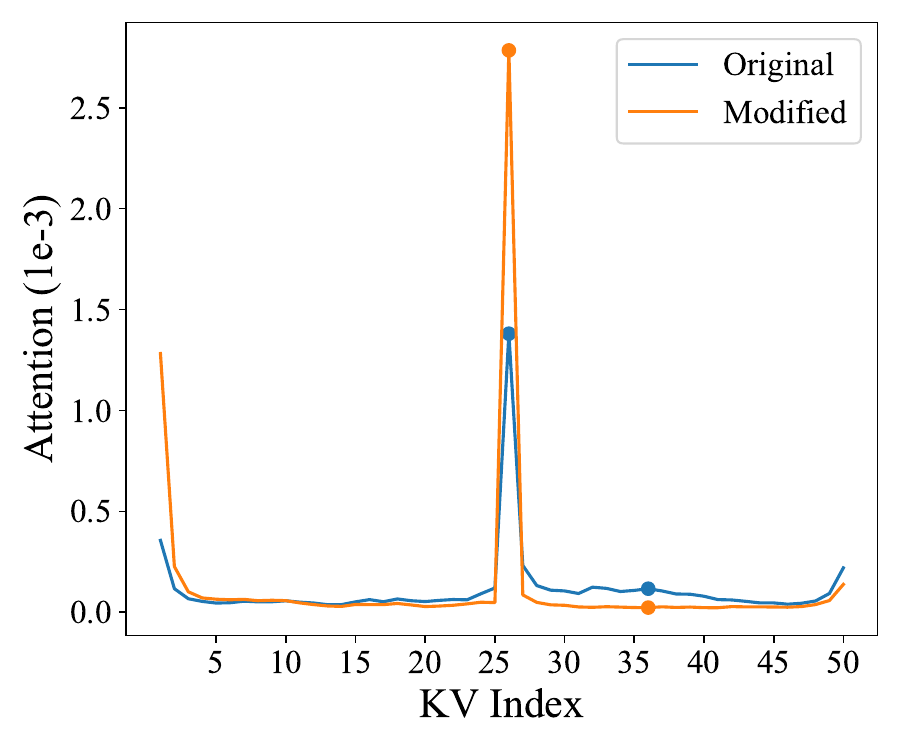}}

  \subfloat[Hidden States]{
    \includegraphics[height=0.7\columnwidth]{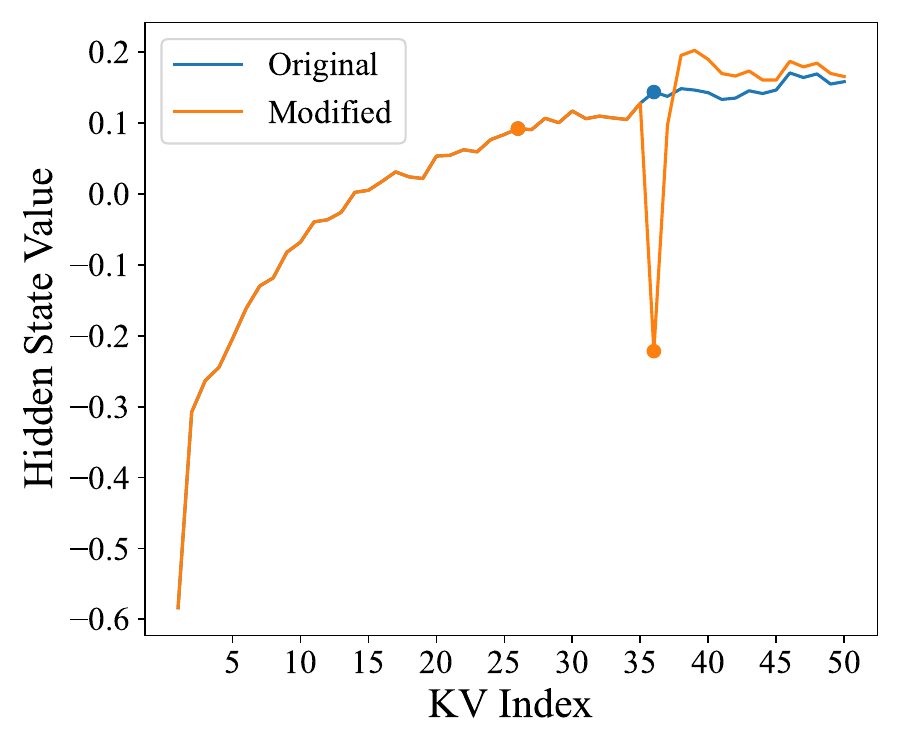}}
  \subfloat[Attention]{
\includegraphics[height=0.7\columnwidth]{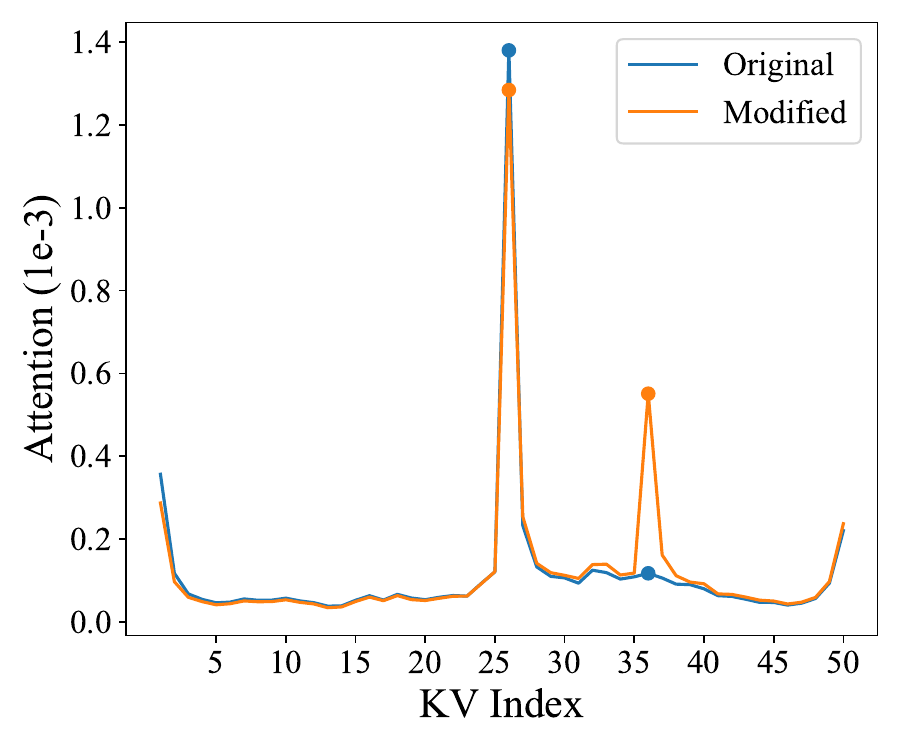}}

  \caption{The positional hidden states and attention of each KV in a KV retrieval task (the gold KV is the 26th KV) of Mistral-7b-v0.2. (a)(b) We modify the hidden states of the 26th KV. (c)(d) We modify the hidden states of the 36th KV. (a)(c) The value of the 213rd channel of the hidden states (averaged across layer 15 to 20) of each KV (averaged across the tokens of the KV). (b)(d) The attention (averaged across layer 15 to 20 and across every attention head) to each KV.
  }
  \label{fig: hidden and attn 26}
\end{figure*}

\subsection{Perturbation on Causal Mask and Position Embedding}
\label{app:Perturbation}
\begin{figure*}[htb]
  \centering
  \subfloat[Original hidden state]{
    \label{sfig: hidden 213}
    \includegraphics[height=0.54\columnwidth]{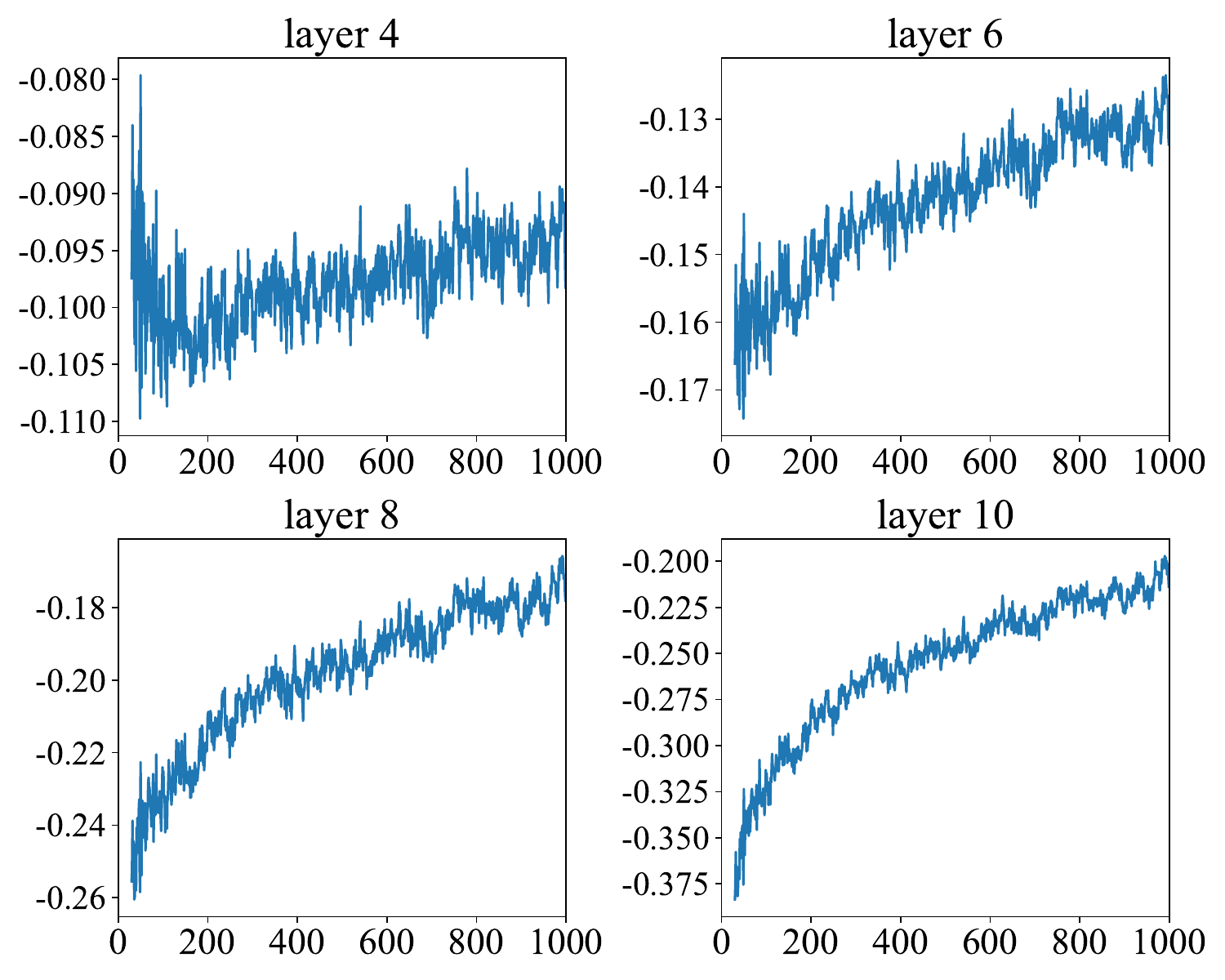}}
  \subfloat[Crop mask]{
\label{sfig: hidden 213 modify mask}
\includegraphics[height=0.54\columnwidth]{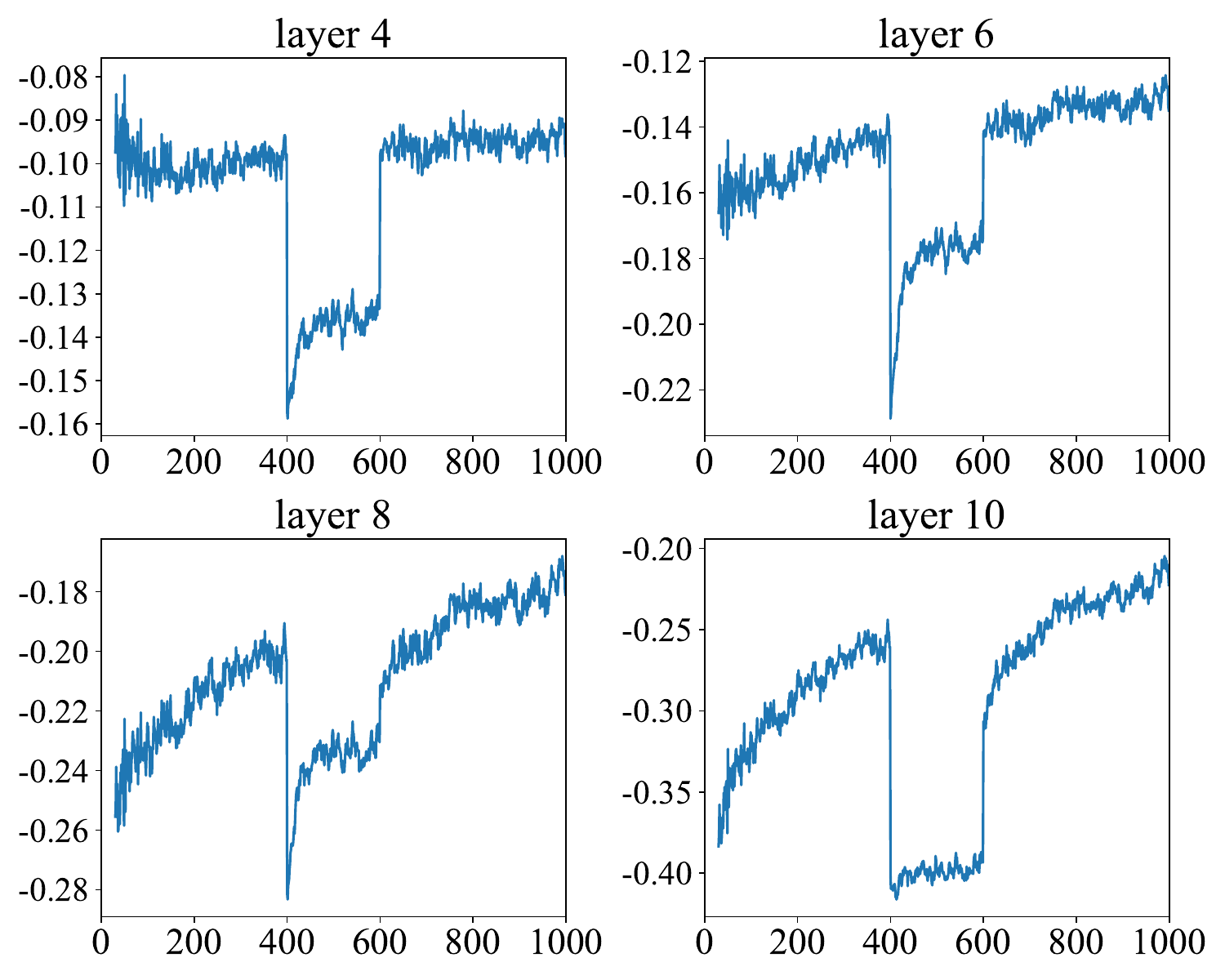}}
  \subfloat[Subtract position ids of PE]{
    \label{sfig:hidden 213 modify PE}
    \includegraphics[height=0.54\columnwidth]{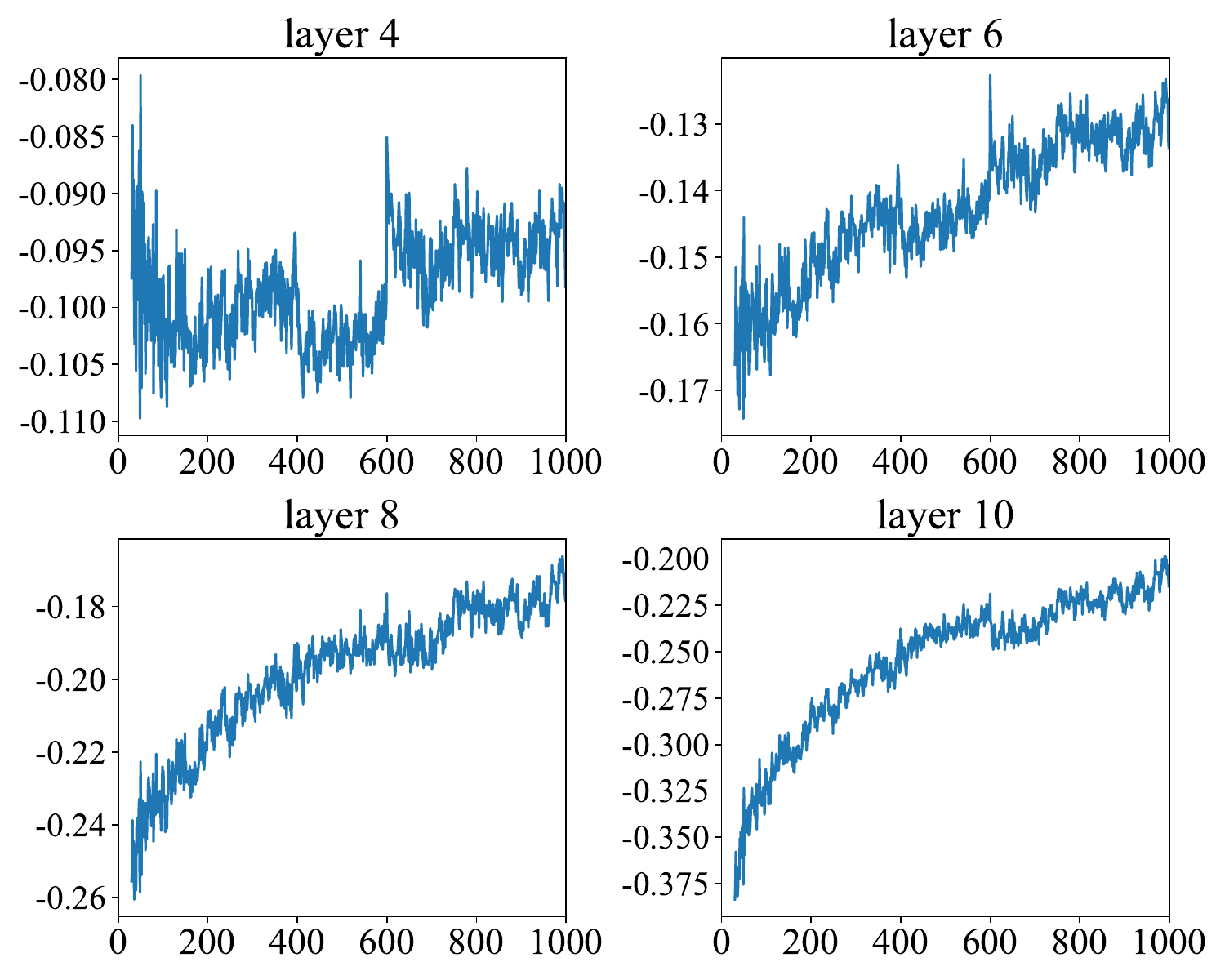}}
  \caption{We performed perturbation experiments on the causal mask and position embedding (PE), showing the 213rd channel of hidden states of some layers of Mistral-7b \citep{jiang2023mistral} using randomly synthesized corpus as input. It is clear that only cropping mask cause significant drop of positional hidden states values.}%
  \label{fig: hidden_change with PE and mask}
\end{figure*}

To further explore the origin of these position hidden states, we performed perturbation experiments. As depicted in Figure~\ref{sfig:hidden 213 modify PE}, subtracting 200 from the position ids corresponding to the 400th to 600th tokens (reducing PE) had only a minor effect on the position hidden states, whereas, in Figure~\ref{sfig: hidden 213 modify mask}, crop the causal mask to make the 400th to 600th tokens unable to attend the 1st to 400th tokens (cropping causal mask) led to significant fluctuations in positional hidden states of the 400th to 600th tokens. This result proves the causal mask is the main factor causing this kind of positional hidden states, and it is the token's position in the causal mask that determines its value in the positional hidden states, but not position ids of position embedding.

\subsection{Positional Hidden States Affect Attention}
\label{app:hidden and attn 26}
Although it is intuitive that these monotonically changing hidden states channels convey rich positional information, it is unclear how much impact one position channel will have on position bias. To find out this, we modify the position hidden states values to see corresponding attention changes, in a KV retrieval task with 50 KV pairs while the gold KV is the 26th KV.

Specifically, we subtract 0.3 from the 213rd channel of the hidden states of the tokens of the 26th and 36th KV respectively (in layer 15 to 20 of Mistral-7b-v0.2), and observe the change of positional hidden states (the 213rd channel) and attention weights of each KV.

As shown in Figure \ref{fig: hidden and attn 26}, we show each KV's average hidden state value of the 213rd channel of Mistral-7b-v0.2 and the average attention to each KV, which are both averaged across layer 15 to 20 (because they are retrieval-related layers). It is clear that when the positional hidden states values of the 26th or 36th KV greatly drop, the attention to it is greatly increased, even over the level of the 1st KV, regardless of whether it is the gold KV. Therefore, we confirm only changing one positional channel can significantly affect attention, thus affect position bias.

\section{Ablation Experiments of the Searching Algorithm}
\label{app:ablation}
\begin{table*}[htb]
	\centering
        \small

 \resizebox{1.7\columnwidth}{!}{
	\begin{tabular}{l|ccccc}
		\toprule
		Method & LLaMA-2-7b & Vicuna-13b & Gemma-7b & Mistral-7b & Qwn1.5-7b\\
            \midrule
            not applied & 31.3  & 50.2 & 28.0 & 59.5 &56.8 \\
            \midrule
            Ours &  40.6 & \textbf{52.7} & \textbf{34.6} & \textbf{60.9} &\textbf{58.3}\\
		w/o monotonicity  & 40.6 & 51.8 & \textbf{34.6}& \textbf{60.9} &\textbf{58.3}\\
  		w/o smoothness  & 40.6 & \textbf{52.7} & 27.8& \textbf{60.9} &\textbf{58.3}\\
  		w/o validation set  & 30.1 &  51.8 &26.5& \textbf{60.9} &\textbf{58.3}\\
            w/ scale top-2 channels &  37.2 & 50.8 &31.7 & 60.1 &57.2\\
            w/ modify last 16 tokens & 41.6  &  51.5  &\textbf{34.6} &59.7 & 58.1\\
            w/ modify all tokens & \textbf{44.0} &50.8  & 31.7 & 59.5 &57.4 \\
            \bottomrule
	\end{tabular}
}
  	\caption{Average performance of different ground-truth positions using different methods on NaturalQuestions multi-document QA dataset (20 documents)~\cite{liu_lost_2023}.}
	\label{tab:ablation_study}
\end{table*}

To evaluate the contributions of different components in our method, we introduce the following sets for the ablation study:
(1) Ours w/o monotonicity, w/o smoothness, and w/o validation set, which adjust the search algorithm by not considering one of these three indicators, respectively (details in Appendix~\ref{subsec:implemention}).
(2) Ours w/ scale 2 channels, which scales the top-2 positional hidden states simultaneously.
(3) Ours w/ modify last 16 tokens and w/ modify all tokens, which adjust the range of tokens affected by the scaling operation in Equ.(\ref{eq:scale_factor}).

Table~\ref{tab:ablation_study} shows the ablation results. It can be seen that without filtering by monotonicity or smoothness, performance may decline, and removing the validation set results in more decline in model performance. When the range of tokens or channels affected by scaling is expanded, most models experience varying degrees of performance loss. Considering these factors, we choose to modify only the last token and the top-1 positional channel to achieve the best performance.

\section{Attention Distribution Layer-wise and Head-wise}
\label{app:Attention Distribution}
\begin{figure*}[htb]
  \centering
\includegraphics[width=2\columnwidth]{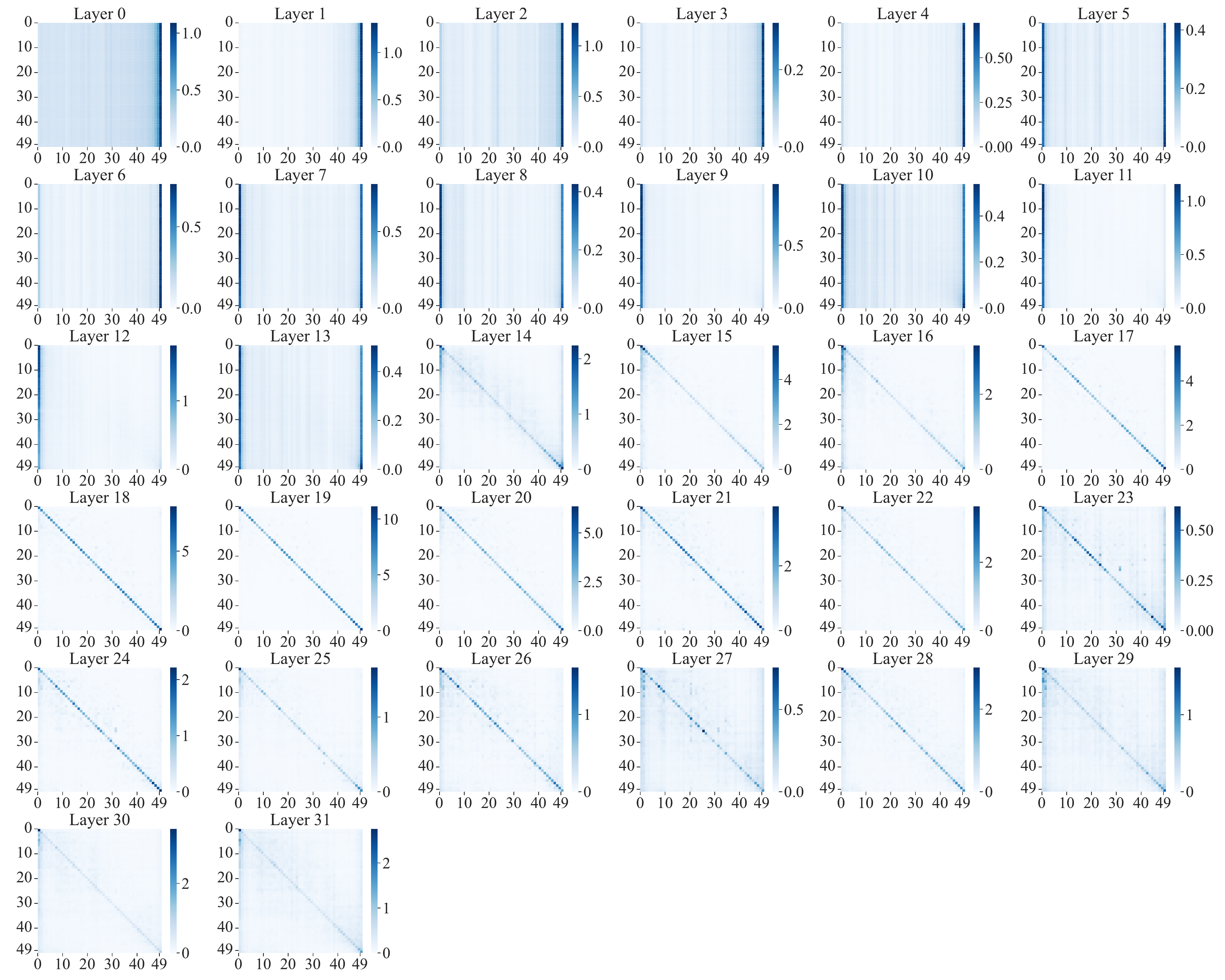}
      \caption{The average attention weight distributed on each KV, of all the 32 layers of Mistral-7b, on a 50 KV pairs retrieval task, when the gold KV is put at each different position.}
        \label{fig: v2last_all_layers_head_mean}
\end{figure*}

Figure \ref{fig: v2last_all_layers_head_mean} shows Mistral-7b's attention to each KV pair of each layer (average across all attention heads) in the context in a KV retrieval task when the gold KV is put at different positions. The y-axis is the gold KV's position, x-axis is each KV's position, and the scale of the colorbar represents attention ($10^{-3}$). We can observe that diagonal patterns, which indicates the attention is concentrated on the ``key tokens'', appear only in the latter layers (start from layer 14), and may be a manifestation of retrieval behavior. In contrast, the former layers only focus on the beginning or end, regardless of where the key information is located.

\begin{figure*}[htb]
    \centering

    \includegraphics[width=2\columnwidth]{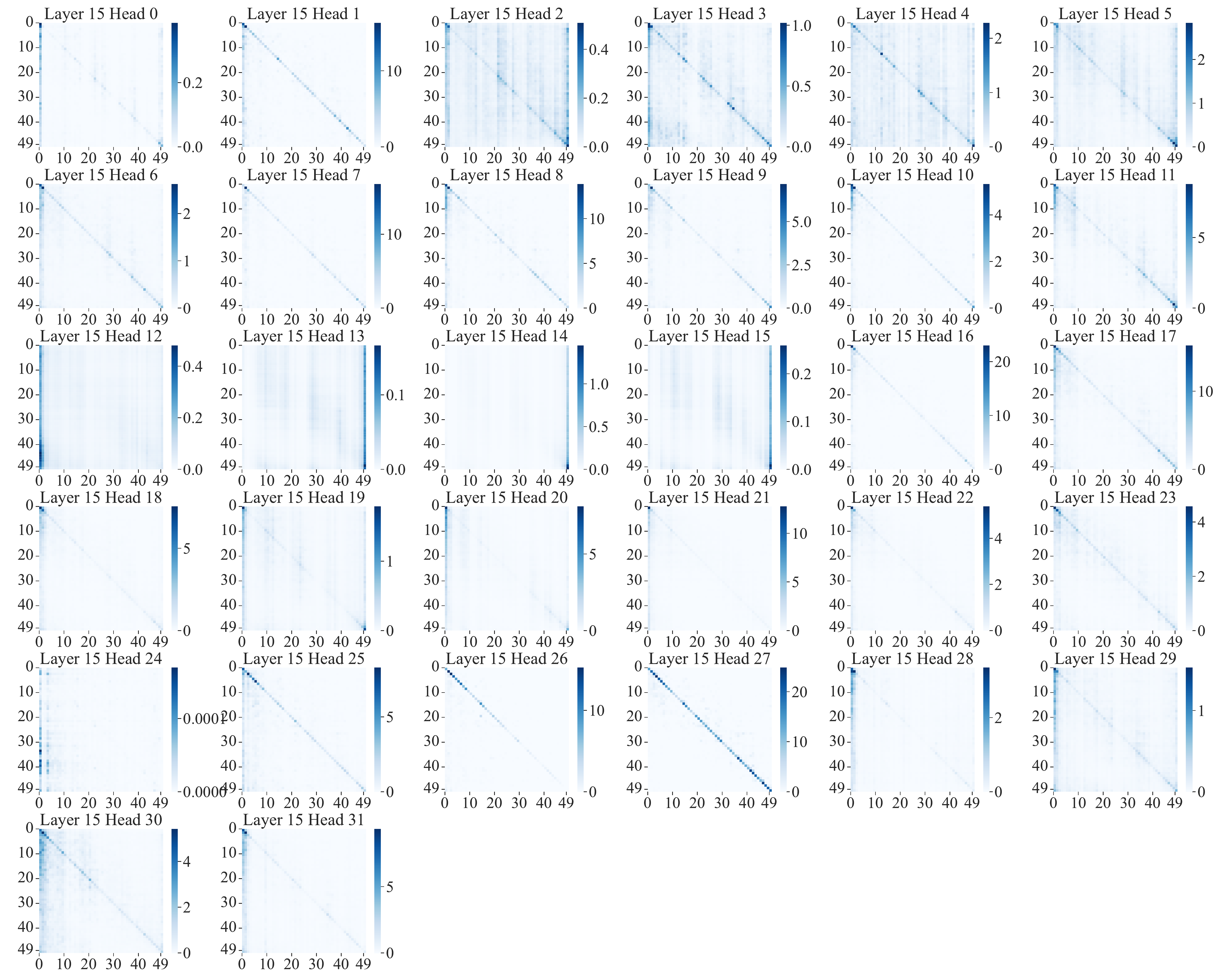}
  \caption{The average attention weight distributed on each KV, of all the 32 attention heads of layer 15 of Mistral-7b, on a 50 KV pairs retrieval task, when the gold KV is put at each different position.}
    \label{fig: v2last_layer15_head_all}
\end{figure*}

Figure \ref{fig: v2last_layer15_head_all} shows the head-wise situation of layer 15. We can see actually only a portion of attention heads exhibit diagonal patterns, which may correspond to \textit{retrieval heads} \citep{wu_retrieval_2024}. The attention distribution in these heads also shows a pattern corresponding ``lost in the middle'', being larger at the beginning or end while significantly smaller at the middle.

\section{Positional Hidden States Visualization}
\label{app:hidden visual}
\begin{figure*}[htb]
  \centering
  \subfloat[mistral-7b]{
\includegraphics[width=1.\columnwidth]{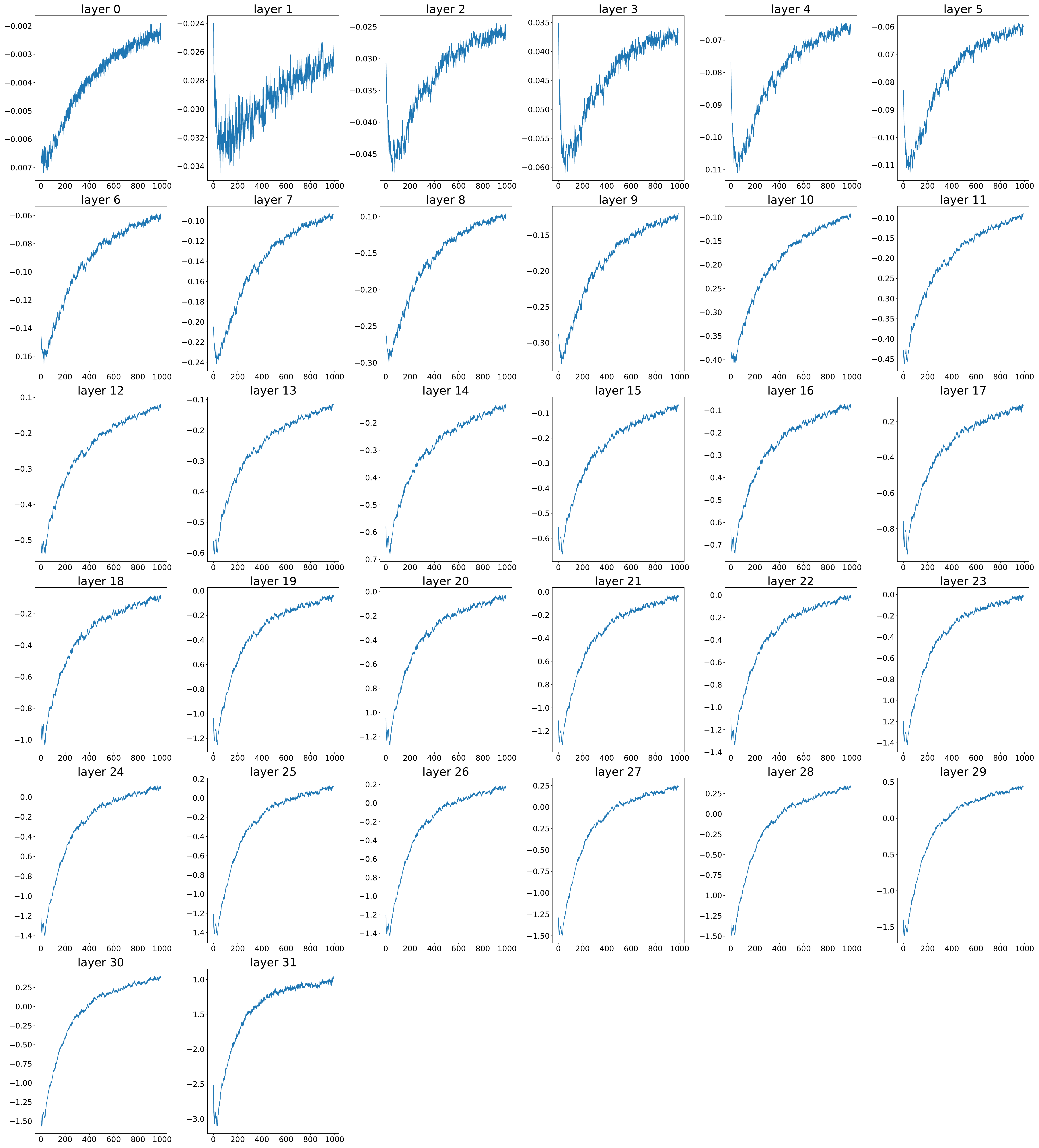}}
\hspace{0.1em}
  \subfloat[LLaMA-2-7b]{
    \includegraphics[width=1\columnwidth]{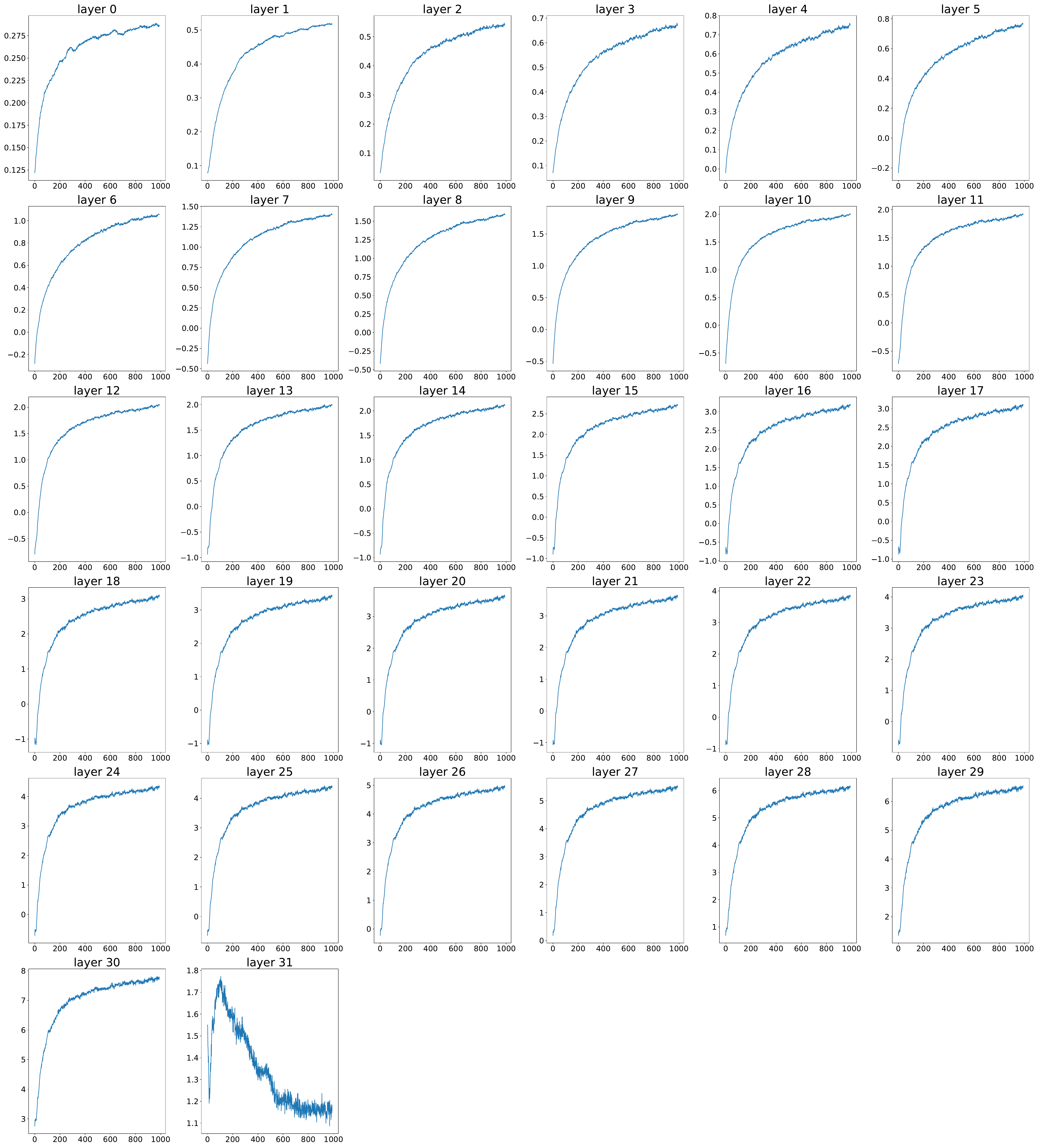}}
    
  \subfloat[MPT-30b]{
\includegraphics[width=1.4\columnwidth]{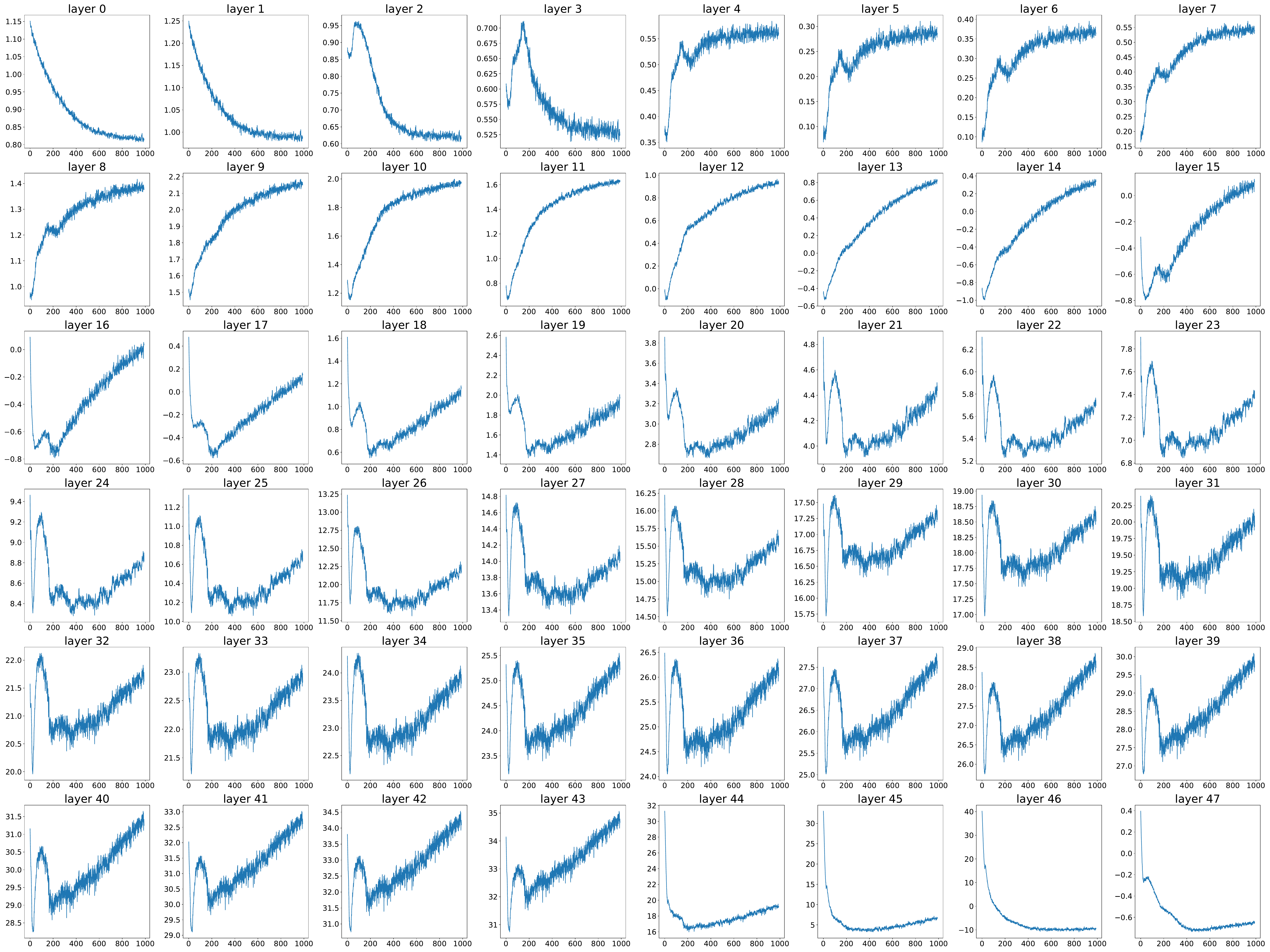}}
\hspace{0.1em}
  \subfloat[Tinyllama-NoPE]{
\includegraphics[width=0.6\columnwidth]{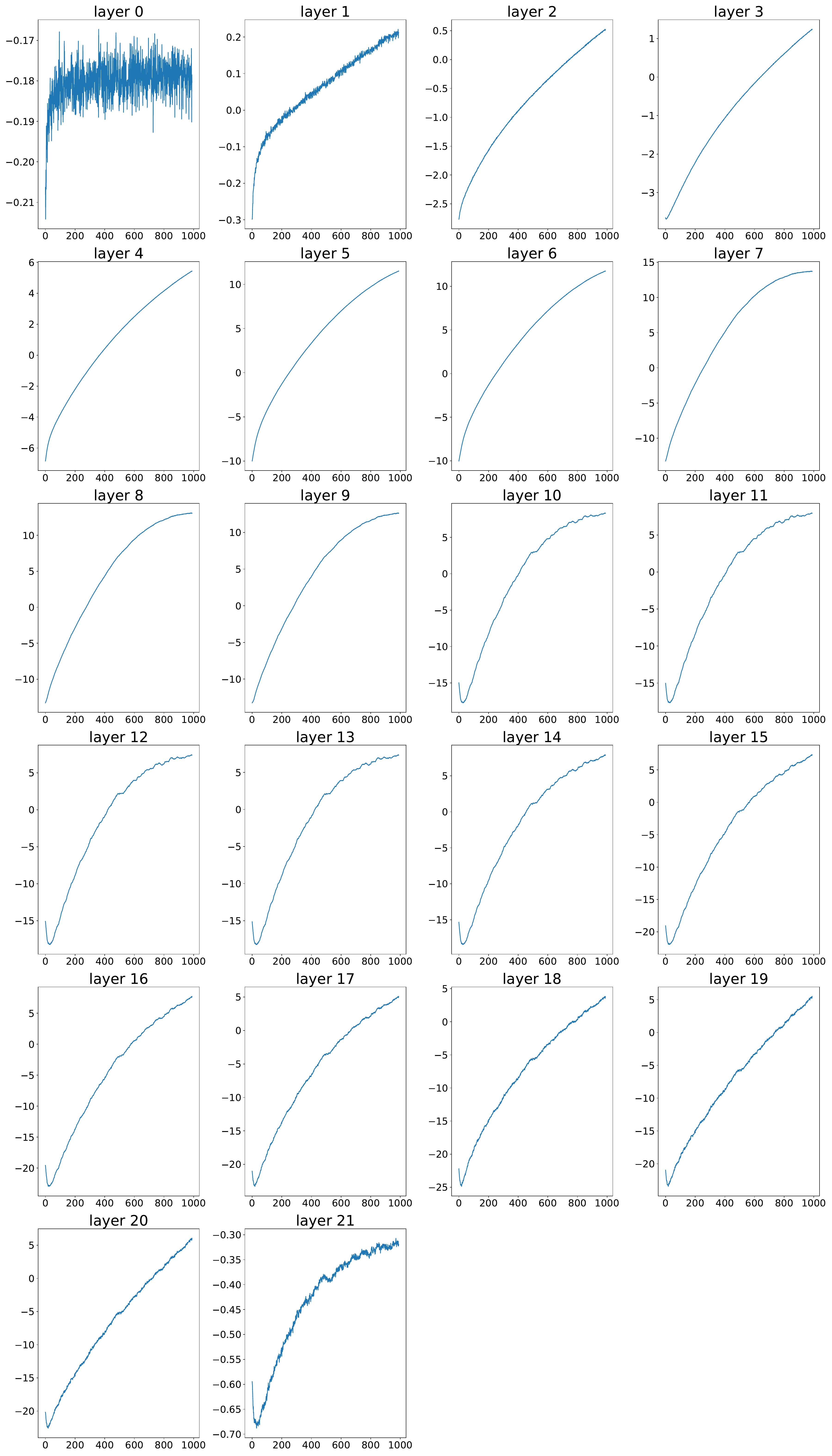}}

  \caption{Hidden states values with the token positions of the positional channel of each layer. The x-axis represents the position, and the y-axis represents the value of the states. (a) The 213rd channel of Mistral-7b-v0.2 (b) The 2393th channel of Llama2-7b (c) The 1942th channel of MPT-30b (d) The 1156th channel of TinyLlama-NoPE-1.1B, which is a model without any position embeddings. }
    \label{fig: hidden layerwise}
\end{figure*}

We show various models' positional hidden states of each layer in Figure \ref{fig: hidden layerwise}. When visualizing, we discard the first 30 tokens because the hidden states values of these tokens are often too huge (usually hundreds of times larger than the normal value \citep{sun_massive_2024}), which can disrupt monotonicity. We observe its monotonic trend may first appears just in the first layer (actually just after the first attention module), and continues to be more marked.

\end{document}